\documentclass[journal]{IEEEtran}
\usepackage{graphicx}
\usepackage{epsfig}
\usepackage{amsfonts}

\usepackage[ruled,vlined]{algorithm2e}
\usepackage{amssymb}
\usepackage{amsmath}
\usepackage{hyperref}
\hypersetup{colorlinks=true, allcolors=blue, urlcolor=purple}
\usepackage{soulutf8}%
\usepackage[colorinlistoftodos]{todonotes}
\usepackage{color}
\usepackage{xargs}%
\usepackage{float}
\usepackage{wrapfig}
\usepackage{multirow}
\usepackage[symbol]{footmisc}
\usepackage{subcaption}
\usepackage{booktabs}
\usepackage{array}
\usepackage{bm}
\usepackage{listings}
\usepackage{xcolor}
\usepackage{bbding}
\usepackage{svg}
\usepackage{enumitem}
\usepackage{tikz}
\usepackage{makecell}

\newcommand{\Circled}[1]{%
  \tikz[baseline=(char.base)]{
    \node[shape=circle, draw, inner sep=1pt] (char) {#1};}}
\usepackage{enumitem} 
\usepackage{tikz} 
\usepackage{graphicx} 

\usepackage{mathtools, stmaryrd}
\usepackage{xparse} \DeclarePairedDelimiterX{\Iintv}[1]{\llbracket}{\rrbracket}{\iintvargs{#1}}
\NewDocumentCommand{\iintvargs}{>{\SplitArgument{1}{,}}m}
{\iintvargsaux#1} %
\NewDocumentCommand{\iintvargsaux}{mm} {#1\mkern1.5mu..\mkern1.5mu#2}
\usepackage{stmaryrd}

\newcolumntype{?}{!{\vrule width 1pt}}
\usepackage{colortbl}
\definecolor{Gray}{gray}{0.85}
\newcolumntype{a}{>{\columncolor{Gray}}c}

\usepackage[numbers]{natbib}

\newcommand{\cut}[1]{}
\usepackage{natbib}

\begin{document}
\title{Enhancing Remote Sensing Vision-Language Models Through MLLM and LLM-Based High-Quality Image-Text Dataset Generation}

\author{Yiguo He,  Junjie Zhu, Yiying Li, Xiaoyu Zhang, Chunping Qiu, Jun Wang, Qiangjuan Huang, Ke Yang
\IEEEcompsocitemizethanks{
\IEEEcompsocthanksitem This work was supported in part by the National Natural Science Foundation of China under Grants 62006241, 62206307, and 42201513.
\IEEEcompsocthanksitem Yiguo He,  Junjie Zhu, Yiying Li, Xiaoyu Zhang, Chunping Qiu, Jun Wang, Qiangjuan Huang, and Ke Yang are with Intelligent Game and Decision Lab (IGDL), Beijing, China (e-mail: yangke13@nudt.edu.cn).
}
}
\markboth{IEEE TRANSACTIONS ON GEOSCIENCE AND REMOTE SENSING}%
{Shell \MakeLowercase{\textit{et al.}}: Bare Demo of IEEEtran.cls for Journals}
\maketitle

\begin{abstract} 
The application of Vision-language foundation models(VLFMs) to remote sensing(RS) imagery has garnered significant attention for its superior capability in various downstream tasks. 
A key challenge lies in the scarcity of high-quality, large-scale, image-text paired training data. Recently, several works introduced extensive image-text datasets for RS and trained their VLFMs. 
However, due to the rudimentary methods used for generating captions, the quality of datasets is suboptimal, requiring larger volumes of training data, while only yielding modest performance improvements.
In this paper, we propose a two-stage method named MpGI(Multi-Perspective Generation and Integration) for generating high-quality text captions for RS images. Firstly, we generate distinct and detailed descriptions from different perspectives using Rule-MLLM(Multimodal Large Language Model) Relay Generation and MLLMs generation methods. Next, we utilize Large Language Models (LLMs) to integrate these diverse descriptions into comprehensive captions, capturing details from multiple perspectives. Finally, We have created the HQRS-IT-210K dataset including about 210,000 RS images and 1.3 million captions. 
We fine-tuned two VLFMs using our dataset: CLIP, a discriminative model, and CoCa, an image-to-text generative model. This process resulted in our proposed HQRS-CLIP and RS-CoCa models.
Experimental results demonstrate that HQRS-CLIP surpassed the previous SOTA RS CLIP model in various downstream tasks while using only 4.2\% of the training data. RS-CoCa outperforms other advanced approaches across benchmark datasets and can generate captions for RS images that rival or even exceed manual annotations. Dataset, pre-trained models, and codes will be released at \url{https://github.com/YiguoHe/HQRS-210K-and-HQRS-CLIP}.
\end{abstract}

\begin{IEEEkeywords}
Remote Sensing, Image-Text Paired Dataset, CLIP, CoCa, Vision Language Foundation Model
\end{IEEEkeywords}

\IEEEpeerreviewmaketitle

\section{Introduction}
\label{sec:intro}

\begin{figure*}[htb]
  \centering
  \includegraphics[width=\linewidth]{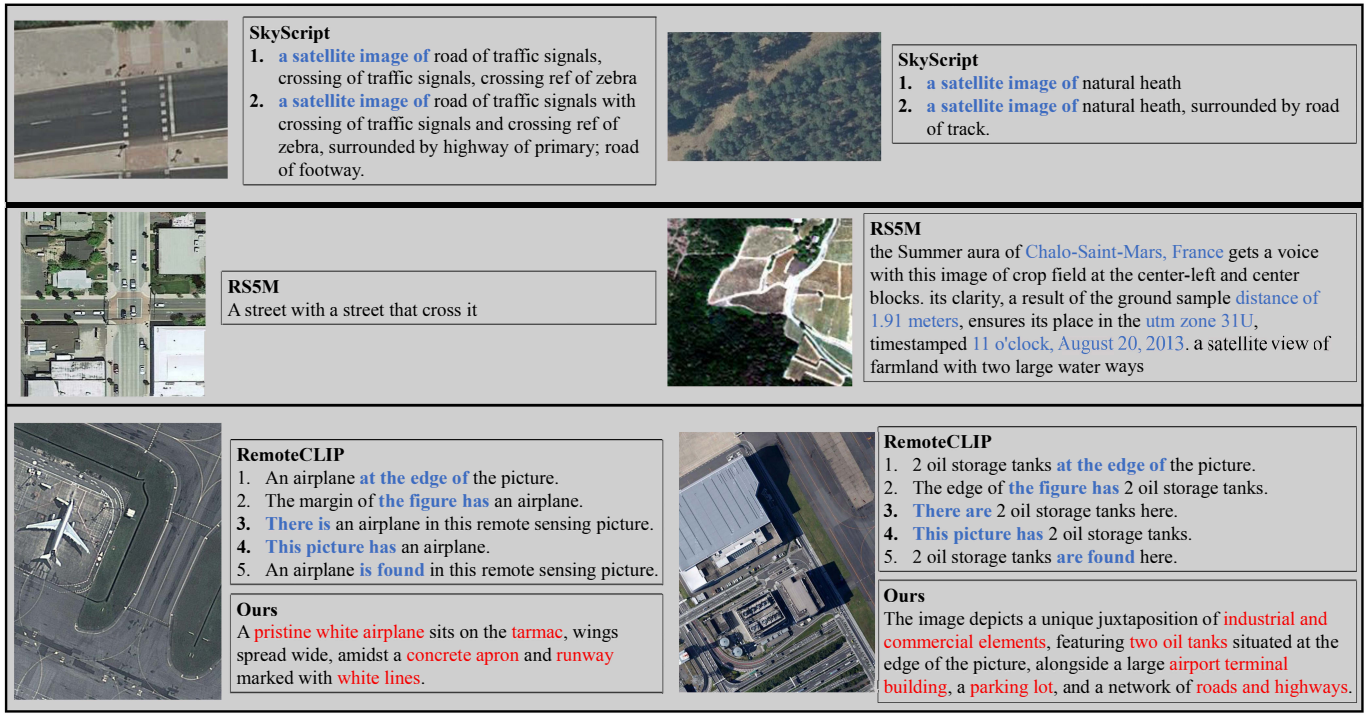}
  \caption{Samples of large-scale RS image-text datasets. SkyScript captions are rule-based, resulting in highly uniform sentence structures and poor alignment with images. In RS5M, BLIP2 captions are short and lack richness, while metadata-based captions are lengthy but misaligned with visuals. RemoteCLIP captions are repetitive and lack diversity. In contrast, our dataset provides accurate, comprehensive, and diverse captions, rivaling human annotations.}
  \label{fig:CaptionCompairation}
\end{figure*}

Vision-language foundation models(VLFMs) bridge visual and textual modalities, enabling comprehensive understanding beyond visual recognition~\citep{clip,ALIGN,DeCLIP,eva-clip,coca}. For example, Contrastive Language-Image Pre-training (CLIP) \citep{clip} uses a contrastive loss to link 400 million images and paired text. Leveraging its strong generalization, CLIP has been applied to diverse tasks like image segmentation \citep{GroupViT}, object detection \citep{VilD}, video understanding \citep{Videoclip}, audio recognition \citep{audioclip}, and 3D point cloud processing \citep{pointclip}. It underpins various multimodal large language models (MLLMs)~\citep{Llava,kosmos2,Qwen} and has also proven useful in applications like data noise filtering \citep{semdedup} and image-text quality assessment \citep{DAC}. In these processes, the pre-trained VLFMs distributes the training cost across all downstream tasks while also providing greater opportunities to scale up the model size~\cite{pathways}. Another typical example of VLFM is Contrastive Captioner (CoCa)~\cite{coca}. It is a generative VLFM which enhances CLIP's contrastive loss with a generative auto-regressive loss, enabling it to learn multimodal image-text representations and excel in image captioning tasks. Leveraging an innovative yet simple design, CoCa achieves competitive performance in generation tasks, outperforming more complex models without the need for special optimization.

Recently, the RS community has recognized the power of VLFMs and has begun exploring its applications in the field of RS~\citep{remoteclip,Skyscript,rs5m,rsclip,cliprs,rs-llava,rsgpt,skyeyegpt}. They demonstrated that VLFMs, trained with extensive image-text paired RS data, perform expressively on various RS applications.
\textit{The key to achieving success in this area lies in the high-quality, large-scale RS image-text paired data.} 
High quality refers to two key aspects. For the image, the dataset should primarily consist of clear aerial or satellite images, while minimizing the inclusion of noisy or irrelevant images~\citep{DataFilteringscalingLaws}. Additionally, images of the same category or scene should be as non-redundant and diverse as possible, ensuring high variability within and across categories~\citep{aid,semdedup}. For the text, it is crucial to maintain a strong correlation between the textual descriptions and the image content. Furthermore, the text should provide rich and detailed descriptions of the image content~\citep{DAC,DCI}, allowing for a high degree of alignment between the image and text. Accurate and comprehensive descriptions are crucial, as captions that focus on only a small part of an image can cause category ambiguity~\citep{patternnet,rsicd}.

However, achieving these goals is highly challenging. Unlike natural images, RS images and their associated text descriptions cannot be effectively sourced from the public internet.
Additionally, manually annotating aerial images requires specialized knowledge and is extremely time-consuming~\citep{aid}. This is more challenging for textual caption annotating, as RS images often lack detailed content, making it difficult even for experts to provide diverse annotations.

To bridge this gap, ~\cite{remoteclip} and ~\cite{Skyscript} utilized rule-based methods to convert annotations and labels into captions. However, as shown in Figure \ref{fig:CaptionCompairation}, these captions tend to provide only broad or incomplete descriptions, lacking detailed information about the images. The singularity of the rules leads to the sentence structure being often too rigid and repetitive, lacking natural expression and diversity. 
~\cite{rs5m} fine-tuned the BLIP-2 model using the RSITMD dataset to generate captions for millions of RS images. However, due to inherent caption limitations in the RSITMD dataset, such as brevity, lack of comprehensiveness, and insufficient diversity, the fine-tuned BLIP2  inevitably inherits these issues, resulting in captions that reflect similar distributional characteristics.
~\cite{chatearthnet} used ChatGPT-3.5 and ChatGPT-4V to generate captions for RS images. However, they offered only land cover types and proportion information to ChatGPT, resulting in descriptions without accurate image information. 

As shown in Figure \ref{fig:CaptionCompairation}, these methods mainly encounter two problems: First, none of them can provide detailed and comprehensive descriptions for RS images, lacking sufficient semantic information. Second, the rule-based method struggles to generate natural and meaningful sentences. \textit{Consequently, despite the large scale of these datasets, the models trained using them achieve only mediocre performance due to the data quality issues as mentioned above. }

\begin{figure}[htb]
  \centering
  \includegraphics[width=\linewidth]{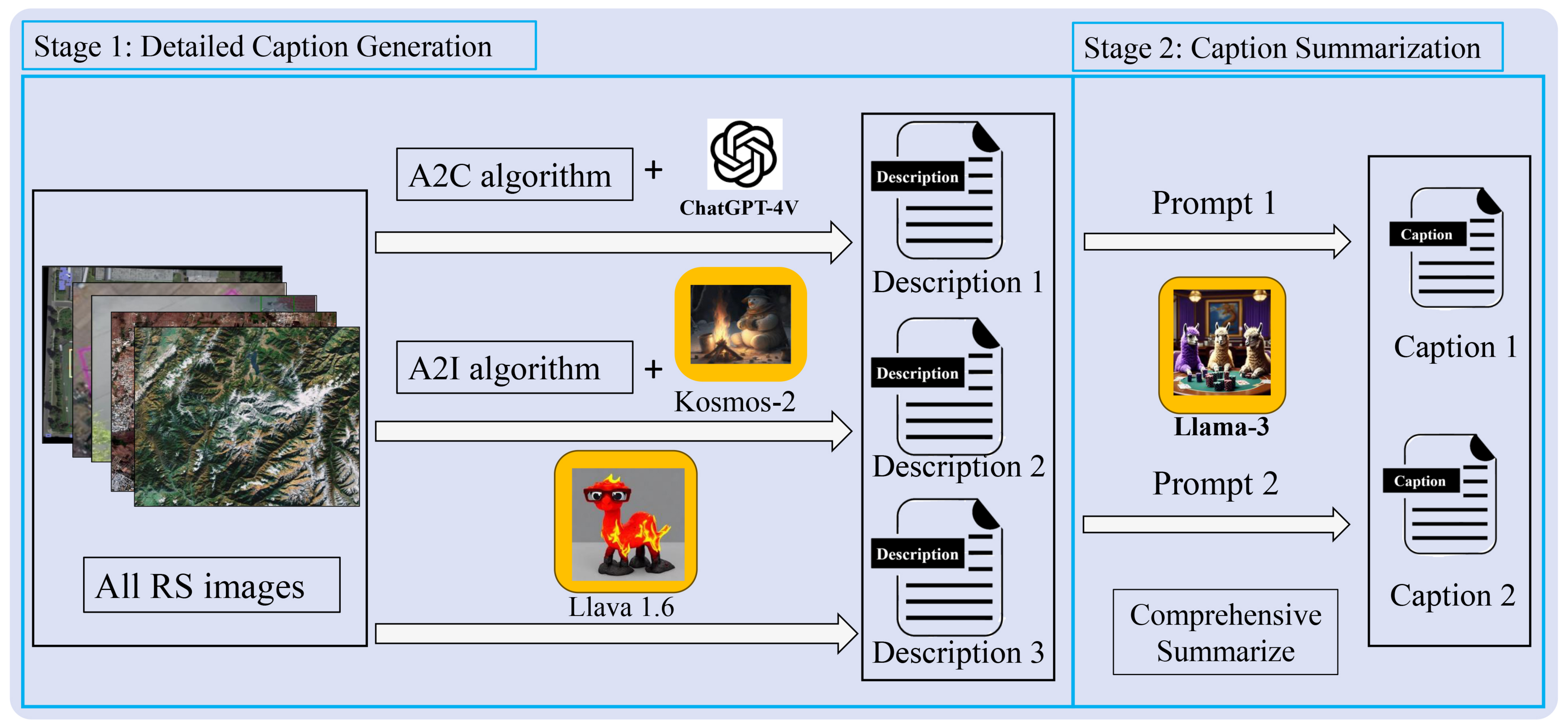}
  \caption{Overview of the MpGI method.}
  \label{fig:pipelineOf2Stages}
\end{figure}

To address the challenges posed by low-quality text descriptions, we proposed a two-stage image-to-text generation method named MpGI(Multi-Perspective Generation and Integration), as illustrated in Figure \ref{fig:pipelineOf2Stages}, to provide more accurate, detailed, and comprehensive descriptions of RS images. 
Firstly, we utilized Rule-MLLM Relay Generation and MLLMs Generation methods to generate accurate and detailed descriptions for each RS image. We provided each image with descriptions from different sources and offered complementary information from various perspectives. At this stage, we generated descriptive content for each image with an average word count exceeding 220 words.
Secondly, we used LLMs to comprehensively summarize the ``multi-view'' descriptions, creating multiple semantically complete image captions with different styles via multiple prompts. This process integrates complementary information from different descriptions, filters out grammatical and semantic errors, and creates captions that capture diverse details from various perspectives. To address the uniform style tendency of LLMs under a single prompt ~\citep{veclip}, we adopted multiple strategies to enhance caption diversity. These include prompting LLMs to randomly sample outputs and using distinct prompt designs to diversify results.

Furthermore, we explored a probability-based fusion method, which allowed the final caption set to include various caption styles generated in the second stage. This can be regarded as a text augmentation strategy without increasing the training budget because it enhanced training data diversity.
Experimental results show that this simple strategy substantially improved model performance without increasing training costs.

Finally, we developed the HQRS-IT-210K dataset (High-Quality RS Image-Text dataset with 210K images), containing approximately 210,000 RS images and 1.26 million image-text pairs. We used only 1/6 data of this dataset to fine-tune CLIP, resulting in our HQRS-CLIP models. Experimental results suggest that our HQRS-CLIP outperformed the previous state-of-the-art methods across various downstream tasks, including zero-shot classification, few-shot classification, RS image-text retrieval, and semantic localization, \textit{all while using only 4.2\% of the training data} (\cite{rs5m}). Then, to validate the effectiveness of our dataset for generation tasks, we fine-tuned the CoCa model using the entire dataset, resulting in RS-CoCa. Experimental results demonstrate that RS-CoCa exhibits significantly improved RS image captioning capabilities, producing captions that can even surpass human annotations in quality.

Our contributions can be summarized as follows.
\begin{itemize}
    \item[$\bullet$] We propose a novel two-stage method named MpGI to generate high-quality RS image-text-paired datasets, taking advantage of advanced LLM and MLLM. After thorough verification, our method yields the HQRS-IT-210K dataset with approximately 210K RS images and 1.26 million image-text pairs. We will release the dataset to advance vision-language research in RS.
    \item[$\bullet$] With the HQRS-IT-210K dataset, we fine-tuned powerful RS VLFMs for both discriminative and generative Vision-Language Tasks. HQRS-CLIP models are capable of extracting robust vision-language representations for various RS applications. RS-CoCa can generate captions comparable or even exceed manual annotations.
    \item[$\bullet$] We conducted extensive ablation experiments to investigate the factors influencing the two stages of the caption generation process, providing valuable guidance for future research. 
    \item[$\bullet$] To compensate the existing image-text retrieval datasets, particularly for evaluating the rapidly evolving capabilities of RS vision-language models (RS VLMs), we proposed the first benchmark dataset and baselines for long-text image retrieval in RS.
\end{itemize}

\section{Related Work}
\subsection{Large-Scale Image-text Datasets Construction} 

\subsubsection{Natural Image-text Datasets Construction}

 \textbf{Manual Annotation.} Manual labeling and crowdsourcing produced datasets like IAPR TC-12~\citep{IAPR-TC-12}, Illinois~\citep{Illinois}, Flickr 8K~\citep{Flickr8K}, Flickr 30K~\citep{Flickr30K}, and MS COCO~\citep{MScoco}. While reliable, this process is costly and insufficient for rapidly expanding data needs.
 
\textbf{Internet Filtering.} Researchers have explored filtering image-text data from the internet to scale data. \cite{ordonez} built the SBU dataset (1M images), and \cite{Conceptualcaptions} created a 3.3M-pair set from billions of pages. Large-scale models like CLIP\citep{clip} (400M), ALIGN\citep{ALIGN} (1.8B), and Florence~\citep{florence} (900M) were trained on such data. However, these predominantly natural-image datasets contain few RS samples, limiting performance in RS tasks~\citep{clip}.

\textbf{Generative Large Models-Based Generation.} With the advent of LLMs and MLLMs, researchers generate captions via models like ChatGPT 3.5 in VIDAL-10M~\citep{languagebind} and ChatGPT-4V~\citep{sharegpt4v} for caption model training. These studies demonstrate the potential of using generative large models to generate image captions for training multimodal models.

\begin{table*}[htbp]
\caption{Comparison of different large-scale RS image-text datasets. \textbf{IT Pairs} refers to image-text pairs. \textbf{Method} refers to the caption generation method. \textbf{Dataset Comment} section analyzes the advantages and limitations of each dataset. \textbf{R-B} refers to Rule-based generation method. \textbf{PNDF} refers to the Public Natural Image-Text Datasets Filtering method.  
} 
\label{tab:Comparison of Datasets}
\centering
\footnotesize
\begin{tabular}{@{}>{\centering\arraybackslash}m{1.5cm} >{\centering\arraybackslash}m{1.2cm} >{\centering\arraybackslash}m{1.2cm} >{\raggedright\arraybackslash}m{12.8cm}@{}}
\toprule
Model & IT Pairs & Method & \multicolumn{1}{c}{Dataset Comment} \\
\midrule
SkyCLIP &  2.6Million & R-B & This approach introduces a large number of new RS images, making a significant contribution. The image labels encompass a wide range of semantic tags, which can enhance the model's generalization ability. \textbf{However}, the caption structure is too rigid and repetitive, lacking both natural expression and diversity. \\
\midrule
RemoteCLIP &  828 K & R-B & By using annotations from public RS datasets, the caption information is accurate. \textbf{However}, the sentence structure of captions is rigid and repetitive, lacking natural expression and diversity. Additionally, the captions often describe only a portion of the image, leading to category ambiguity. Furthermore, the captions lack image background information and scene context beyond the annotations. \\
\midrule
GeoRSCLIP &  5 Million & R-B PNDF MLLM-B & Using fine-tuned BLIP-2 combined with annotated RS labels, it generates captions with natural and varied structures. The dataset includes a large and diverse collection of images, enhancing the model's generalization ability. \textbf{However}, the images obtained by PNDF often include irrelevant or suboptimal ones, and many captions contain metadata unrelated to the visual content, which can distract the model from focusing on the visual information. Additionally, BLIP-2 can only utilize class labels when generating captions, resulting in brief descriptions that lack detailed information about the images. \\
\midrule
Ours &  1.3Million & MpGI & The MpGI method is composed of Rule-MLLM Relay generation, MLLM-Based, and LLM-Based methods. Each image's description is sourced from multiple origins, making it comprehensive and precise. By compressing the Visual information to captions within 77 tokens, the caption information can be fully used by CLIP. Diverse captions capture a wide range of semantic and visual information. \textbf{However}, Although keyword filtering and word-length filtering were applied to reduce noise, inaccuracies caused by hallucinations during the generation process of MLLMs and LLMs cannot be entirely eliminated. \\
\bottomrule
\end{tabular}
\end{table*}

\subsubsection{RS Image-text Datasets Construction}  

\textbf{Manual Annotation.} Annotating aerial images requires specialized knowledge and is extremely time-consuming~\citep{aid}. Datasets such as UCMcaption (with 2,100 images), Sydney Captions (with 613 images)\citep{UCM}, RSICD (with 10,921 images)\citep{rsicd}, RSITMD (with 4,743 images)\citep{rsitmd}, and NWPU-Captions (with 31,500 images)\citep{NWPU-captions} rely on manual labeling. While the captions are accurate, the process is costly and inefficient, which limits the scalability of datasets.

\textbf{Public Natural Image-Text Datasets Filtering.} Unlike natural images, RS image-text data is scarce online. ~\cite{LAION-EO} used CLIP to extracted 28,572 RS images from LAION-EO, but its utility remains unproven. ~\cite{rs5m} filtered 3M RS image-text pairs from 11 datasets, but data quality remained noisy.

\textbf{Rule-based Generation Methods.} Rule-based approaches generate RS captions by assembling existing annotations or metadata of RS images~\citep{remoteclip,Skyscript,rs5m}. However, these captions often suffer from high repetition, lack natural semantics, and contain irrelevant metadata noise that is difficult to align with visual features in the images.

\textbf{Generative Large Models-Based Generation Methods.} Recently, researchers have begun using LLMs and MLLMs to generate RS captions.~\cite{chatearthnet} employed ChatGPT 3.5 and 4V for RS captioning, but results were limited by simple prompts and lack of detail. ~\cite{rs5m} fine-tuned the BLIP2 model using the RSITMD dataset to captioning millions of RS images. However, current methods fail to fully leverage existing LLMs and MLLMs. Chatearthnet did not effectively provide useful information to ChatGPT, while the tuned BLIP2 used by RS5M inherits the inherent limitations of RSITMD. As a result, these methods have not fully harnessed the capabilities of generative large models. 

As shown in Table \ref{tab:Comparison of Datasets}, we compare the large-scale training data used by various existing RS CLIP models. We can gain some valuable takeaways from it. Firstly, leveraging publicly available datasets to construct RS image-text datasets is a common practice today, primarily because it is cost-effective and leverages previously accurate manual annotations. Secondly, some studies are starting to focus on using large models for labeling RS images. This approach effectively harnesses the benefits of large-scale pre-trained models in the natural image domain to generate natural sentences. Additionally, large models can utilize existing RS dataset annotations, resulting in highly accurate captions.

\subsection{Foundation Models and RS VLFMs}

\textbf{Foundation Models.} Foundation Models (FMs) surged with the emergence of transformer~\citep{Transformer}, trained on broad data and can be adapted to a variety of downstream tasks~\citep{FMs}. 
In natural language processing (NLP), LLMs like BERT~\citep{bert}, T5~\citep{T5}, BART~\citep{bart}, and the GPT series~\citep{GPT1,GPT2,GPT3,GPT4} achieved significant success. Decoder-only models like GPT-3 are pre-trained on next-token prediction tasks, demonstrating strong zero-shot and few-shot generalization.

\textbf{Vision-Language Foundation Models.} In computer vision (CV), traditional pretraining methods for large models relied on manually labeled datasets~\citep{clip}, which confined models to predefined object categories and necessitated additional labeled data to recognize new concepts. To bring zero-shot capabilities similar to GPT-3 into CV, ~\cite{clip} employed a contrastive learning approach, training CLIP on 400 million image-text pairsCLIP is highly competitive in zero-shot transfer tasks, performing on par with task-specific supervised models. This work inspired the development of subsequent models such as ALIGN~\citep{ALIGN}, DeCLIP~\citep{DeCLIP}, and EVA-CLIP~\citep{eva-clip}.

\textbf{RS VLFMs.} Although CLIP has shown impressive performance, it underperforms in specialized fields such as medicine and RS~\citep{clip,rs5m}. In these contexts, developing domain-specific CLIP models can offer substantial advantages. For instance, in the medical domain, MedCLIP~\citep{medclip} fine-tuned CLIP using only 20K medical image-text pairs and surpassed previous state-of-the-art methods. Similarly, researchers in RS are developing specialized foundational models based on CLIP, such as RemoteCLIP~\citep{remoteclip}, SkyCLIP~\citep{Skyscript}, and GeoRSCLIP~\citep{rs5m}.

\textbf{RS MLLMs.} Integrating CLIP’s image encoders with LLMs yields MLLMs (e.g., BLIP~\citep{blip2,instructblip}, Kosmos~\citep{Kosmos1,kosmos2}, Llava~\citep{Llava,llava15,llava16}, Qwen-VL~\citep{Qwen}). 
In RS, BITA~\citep{BITA} builds upon pre-trained large models by further training them on RS datasets for bridging the image-text modality gap. RSGPT~\citep{rsgpt} and RS-LLaVA~\citep{rs-llava} were developed by fine-tuning InstructBLIP and LLaVA on limited datasets like RSICap and RS-instructions, which involve complex data construction method and remain small in scale. The SkyEye-968k dataset~\citep{skyeyegpt}, constructed by integrating existing captioned datasets and manually verified samples, includes 968k samples but heavily relies on existing captioned data, making it costly and less diverse. 

In contrast, our method prioritizes strong image-text alignment, enabling efficient scalability to millions of samples using simple methods. These aligned data are well-suited for VLFMs like CLIP and CoCa, which can be trained efficiently with such simple formats.

\section{High-Quality Image-Text Pairs Generation}
\label{Dataset Construction}
We collected 210,556 RS images from 23 public satellite and UAV (unmanned aerial vehicle) imagery datasets and captioned them using a two-stage framework, as shown in Figure.  \ref{fig:pipelineOf2Stages}.

\begin{table*}[htbp]
\caption{Overview of HQRS-IT-210K Dataset. For the fMoW dataset, we only selected samples from the validation set, as GeoRSCLIP~\citep{rs5m} had already generated captions for the training set, thereby avoiding redundancy.}
\label{tab:Overview of HQRS-IT-210K}
\centering
\footnotesize 
\begin{tabular}{c c c c p{12cm}} 
\toprule
Collection & Datasets & Images & Class & \multicolumn{1}{c}{Description} \\ 
\midrule
\multirow{1}[3]{*}{MSrgb*1} & \multirow{1}[3]{*}{fMoW} & \multirow{1}[3]{*}{22723} & \multirow{1}[3]{*}{62} & The fMoW dataset includes class and bounding box annotations with pan-sharpened RGB images derived from multispectral data to improve resolution and visual detail. \\ 
\midrule
\multirow{8}{*}{CLS*8} & AID & 5465 & 30 & \multirow{8}{12cm}{In RS classification datasets, images within the same category exhibit high similarity and significant homogeneity. As a result, during the P-Hash deduplication process, many samples were excluded from the original dataset. These RS scene classification datasets only contain class labels, so during rule-based caption generation, GPT-4 is also used to expand the sentences for each category's caption by identifying common features across multiple images of the same category within each dataset. } \\ 

 & RESISC45 & 13630 & 45 & \\ 
 & OPTIMAL & 1596 & 31 & \\ 
 & RSI-CB256 & 9582 & 35 & \\ 
 & RSI-CB128 & 13184 & 45 & \\ 
 & EuroSAT & 5101 & 10 & \\ 
 & WHURS19 & 846 & 19 & \\ 
 & MLRSNet & 22975 & 46 & \\ 
\midrule
\multirow{4}{*}{UAV*4} & VisDrone & 27303 & 11 & \multirow{4}{12cm}{To increase data diversity, we introduced Drone aerial imagery datasets, which include detailed class and bounding box annotations. The ablation experiment results indicate that the image-text data constructed from these datasets can enhance the recognition capability of the CLIP model in the RS domain.} \\ 
 & Stanford & 3067 & 6 & \\ 
 & AUAIR & 2175 & 8 & \\ 
 & CARPK & 513 & 1 & \\ 
\midrule
\multirow{4}{*}{Seg*4} & iSAID & 20786 & 15 & \multirow{4}{12cm}{These datasets provide more comprehensive visual information, such as the categories of all objects in an image, thereby offering annotations and textual information with higher alignment accuracy. To ensure consistent processing, we convert segmentation masks to bbox format and process them together with the object detection dataset in subsequent stages.} \\ 
 & LoveDA & 4187 & 6 & \\ 
 & Vaihingen & 33 & 5 & \\ 
 & Potsdam & 4515 & 5 & \\ 
\midrule
\multirow{6}{*}{DET*6} & HRRSD & 21761 & 13 & \multirow{6}{12cm}{These RS object detection datasets provide precise class and bounding box annotations, allowing rule-based generation methods to produce relatively informative captions. However, since these datasets are designed for object detection tasks, many elements that are not emphasized in object detection, such as image scenes, remain unannotated. Nonetheless, textual descriptions of these image features are crucial for image-text alignment tasks. The extraction of such information requires the use of MLLMs.} \\ 
 & LEVIR & 3791 & 3 & \\ 
 & RSOD & 936 & 4 & \\ 
 & DOTA & 1869 & 15 & \\ 
 & DIOR & 23463 & 20 & \\ 
 & HRSC & 1055 & 31 & \\ 
\bottomrule
\end{tabular}
\end{table*}

\subsection{Data Preparation}
\begin{figure*}[htb]
  \centering
  \includegraphics[width=\linewidth]{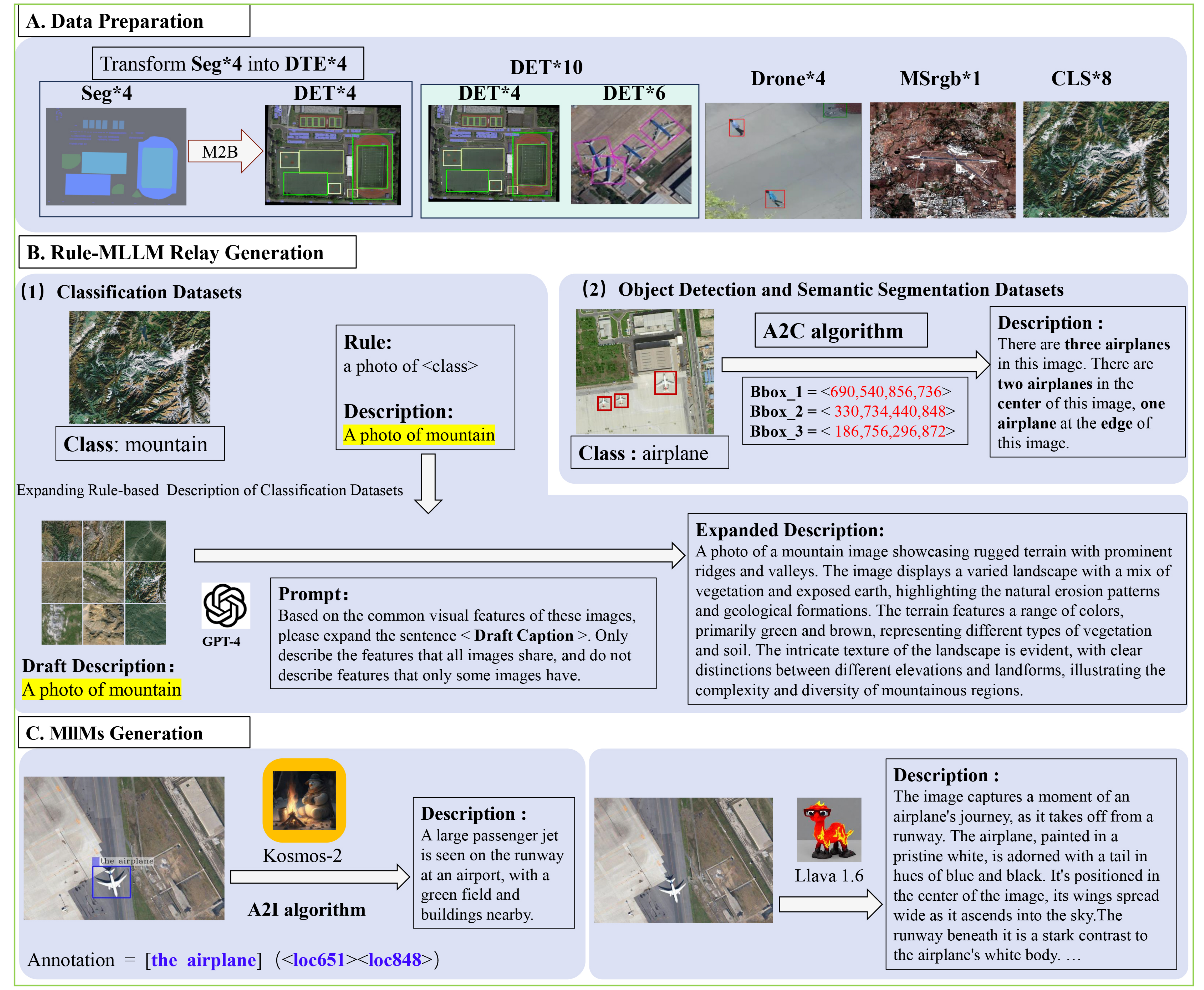}
  \caption{Overview of the caption generation method. A. After being transformed into DET*4 by the M2B algorithm~\citep{remoteclip}, Seg*4 combined with DET*6 to form DET-10. }
  \label{fig:pipelineOfStage1}
\end{figure*}

As shown in Figure \ref{fig:pipelineOfStage1} A, we collected 23 datasets and thoroughly cleaned the images and annotations before generating captions. In Table \ref{tab:Overview of HQRS-IT-210K}, we provide a comprehensive overview of the sources of the HQRS-IT-210K dataset, detailing the information for each sub-dataset.

In terms of images, we first removed unannotated images from each dataset. For images that are too large, we employ a sliding window approach to partition them into several non-overlapping smaller image patches. This is to capture the fine-grained details of these large RS images when generating captions, while also serving as a data augmentation strategy. A strict deduplication method using p-hash and URLs has been employed to prevent data leakage~\citep{remoteclip}. Ultimately, about \( \frac{1}{5} \) images were removed. In terms of annotations, denoising was also performed. For example, we removed annotations labeled as “ignored regions” and “others” in the VisDrone dataset. The wording of the original annotations was adjusted where necessary. For example, we changed the annotation for people from “Human” to “person”.

\subsection{Stage 1: Detailed Caption Generation} \label{sec 1: stage 1}

As shown in Figure \ref{fig:pipelineOfStage1} B and C, our goal is to generate detailed and accurate descriptions for each RS image by various methods. Each method has its advantages, offering complementary perspectives in image descriptions. When combined, these descriptions provide accurate and enriched information for each image.

\subsubsection{Rule-MLLM Relay Generation}
Our images are from classification, object detection, and segmentation datasets. They have been precisely annotated with accurate class labels, bounding boxes, and segmentation masks. Our goal is to efficiently use these precise annotations.

For segmentation masks, we use Mask-to-Box (M2B) algorithm~\citep{remoteclip} to extract the coordinates (xmin, ymin, xmax, ymax) of objects annotated in the Seg*4 into bounding boxes, converting them into DET*4. DET*4 and DET*6 were combined into DET-10 for unified caption generation. As shown in Figure \ref{fig:pipelineOfStage1} B (2), for DET-10, we use the Annotation to Description (A2D) algorithm to convert the class labels, object counts, and location annotations into English sentences. When describing the positions of objects, we define the central area as the rectangular region spanning from 1/4 to 3/4 of the image's width and height, with the remaining area defined as the edge area. The A2D algorithm improves upon the B2C algorithm ~\citep{remoteclip} by reducing category ambiguity.

The A2D algorithm includes the following rules:

\begin{itemize}
    \item[$\bullet$] \textbf{Rule 1}: Describe all objects annotated in the image.
    
    ~~ \textit{Example: There are three cars and two trucks in this image.}
    
    \item[$\bullet$] \textbf{Rule 2}: Describe objects located both in the center and at the edge of this image.
    
    ~~ \textit{Example: There are three cars in the center of this image and two trucks at the edge of this image.}
    
\end{itemize}

The pseudo-code for the A2D algorithm is shown in Algorithm \ref{alg:a2d}.
These rules ensure that all annotated objects are included in the captions, alleviate the risk of category ambiguity~\citep{patternnet}, and provide a more complete description.

\begin{algorithm}
\caption{Annotation to Description (A2D)}
\label{alg:a2d}
\SetAlgoNlRelativeSize{-1} 
\SetKwFunction{FMain}{Annotation\_to\_Description}
\SetKwProg{Fn}{Function}{:}{\KwRet{descriptions}}

\Fn{\FMain{bbox\_list}}{
    descriptions $\gets$ [ ]\;
    \tcp{\textbf{Rule 1}: Describe objects annotated in the image.}
    class\_counts $\gets$ count\_categories(bbox\_list)\;
    description\_parts $\gets$ generate\_description\_parts(class\_counts)\;
    descriptions.append("There are " + join(description\_parts, ", ") + ".")\;
    \tcp{\textbf{Rule 2}: Describe objects located both in the center and at the edges of the image.}
    center\_counts, edge\_counts $\gets$ count\_center\_and\_edge(bbox\_list)\;
    center\_parts $\gets$ generate\_description\_parts(center\_counts)\;
    edge\_parts $\gets$ generate\_description\_parts(edge\_counts)\;
    descriptions.append("There are " + join(center\_parts, ", ") + " in the center, and " + join(edge\_parts, ", ") + " at the edges of the image.")\;
}
\end{algorithm}

As shown in Figure \ref{fig:pipelineOfStage1} B (1), for classification datasets, we start by generating the original descriptive sentence with “a photo of [class]”. Then We use ChatGPT-4V~\citep{GPT-4v} to expand this sentence based on the common features of images in the same class folder. We chose ChatGPT-4V for its exceptional performance. Since each image category requires only one expansion, the cost remains low despite ChatGPT-4V being closed-source. Specifically, for each class, we randomly select 9 images from the class folder and pass them to ChatGPT-4V, which identifies their common features to expand the original descriptive sentence. We refer to this approach (first generating simple captions using rule-based methods and then expanding them with ChatGPT—4V) as the "Rule-MLLM Relay Generation method", abbreviated as "R-M Relay". Experimental results demonstrate the effectiveness of these strategies (See Figure \ref{table:AblationOfStrategies}).

To address the hallucination issue, we manually conducted a sentence-by-sentence review of all captions generated by ChatGPT-4V to ensure the accuracy of the caption content.

\subsubsection{Instruction-Guided MLLMs Generation} 

The descriptions of DET-10 generated by the rule-based method lack contextual information beyond annotations, such as details about aerial scenes. The descriptions of the classification datasets lack individual details of each image. So we leverage MLLMs to generate more detailed and specific descriptions for images.

First, we selected the Kosmos-2~\citep{kosmos2} model to generate fine-grained information. The selection was motivated by two key reasons: (1) Kosmos-2 is one of the most advanced open-source VLMFs, equipped with visual grounding capabilities and achieving state-of-the-art performance in image captioning tasks; (2) it has the ability to accept object detection boxes as inputs, enabling it to generate directed and targeted captions. This capability aligns perfectly with the annotation requirements of our DET-10 dataset for making use of their bbox annotations.

Then, We also chose Llava-1.6~\citep{llava16} (also called LLaVA-NeXT) to generate detailed descriptions for each image. Firstly, Llava 1.6 and Kosmos-2 differ in model architecture and training data, leading to variations in their captioning abilities. These two models can generate descriptions from different perspectives for the same image, often providing complementary information. Secondly, the image resolution of Kosmos-2 is set to 224×224, which leads to the loss of contextual details in many large RS images. So we need another MLLM capable of handling high-resolution images to generate detailed descriptions for each image. Llava-1.6 supports resolutions up to 672x672, 336x1344, and 1344x336 across 3 aspect ratios, enabling the capture of more visual details. Moreover, Llava-1.6 demonstrates exceptional performance, outperforming other open-source MLLMs such as instructBLIP~\citep{instructblip}, CogVLM~\citep{cogvlm}, and Yi-VL~\citep{yi-vl}.

The instruction format and prompt have significant impacts on the performance of MLLMs. To achieve better results, we explored the impact of different instructions for Kosmos-2 and Llava-1.6 before formal experiments. We randomly selected 1,000 images from the overall dataset to test the impact of different instructions.

For Llava-1.6, we primarily tested its ability to generate detailed descriptions within instructions including class labels or not. Experimental results showed that the instruction \textit{Describe this image in detail} can guide Llava-1.6 to generate accurate and detailed information for RS images.

For Kosmos-2, after careful designation and extensive experiments with 10 different instruction templates, we adopted the following 3 instruction templates:

\begin{enumerate}[label=\protect\Circled{\arabic*}, itemjoin={{, }}, itemjoin*={{, and }}]
    \item[\Circled{\scriptsize 1}] Describe this image with [class] in detail:
    \item[\Circled{\scriptsize 2}] Describe this image with [class + bbox] in detail:
    \item[\Circled{\scriptsize 3}] Where is/are the [class + bbox]? Answer:
\end{enumerate}

For images with only class label, we use \Circled{\scriptsize 1} to generate accurate and comprehensive descriptions. For images with one or two bounding boxes, we use both \Circled{\scriptsize 2} and \Circled{\scriptsize 3} to obtain accurate positional information and comprehensive image descriptions. For images with more than two annotated bboxes, we use \Circled{\scriptsize 1} to achieve comprehensive image descriptions. 

Since Kosmos-2 can not directly accept bbox or semantic segmentation mask as input, we propose the Annotation-to-Instruction (A2I) algorithm. It automatically converts class labels, object detection bounding boxes, and segmentation masks into instructions that Kosmos-2 can perceive.  The pseudo-code is shown in Algorithm \ref{alg:a2i}. These instructions, along with the images, are then fed into Kosmos-2 to generate captions. 

We used carefully designed instructions to guide MLLMs in generating accurate and complementary descriptions for RS images. Therefore, we refer to this approach as Instruction-Guided MLLM generation. Ultimately, we obtained detailed captions with an average of approximately 220 words for each RS image. 

\textbf{Hallucination Phenomena and Countermeasures.} Generative large models inevitably encounter hallucination issues~\cite{hallucination} in the process of text generation. To analyze these problems, we randomly sampled 200 captions and identified three main types of hallucinations: (1) Parametric Knowledge Bias: For example, the model misinterprets a generic cityscape as a specific city and appends incorrect temporal or locational details; (2)Intrinsic Hallucination: For example, instances include generating garbled text or misidentifying wind turbines in an image as helicopters; (3) Misjudgment due to Safety Mechanisms: The model incorrectly outputs refusals such as "content does not comply with regulations". 

To address these issues, we implemented the following strategies: (1) Prompt design was applied in subsequent stages to guide the LLM to ignore non-visual information, such as time and location; (2) Regular expressions were employed to filter out garbled text and invalid sentence structures. While these strategies effectively resolved the majority of issues, further research is required to mitigate hallucinations related to the misinterpretation of image content.

\begin{figure}[htb]
  \centering
  \includegraphics[width=0.48\textwidth]{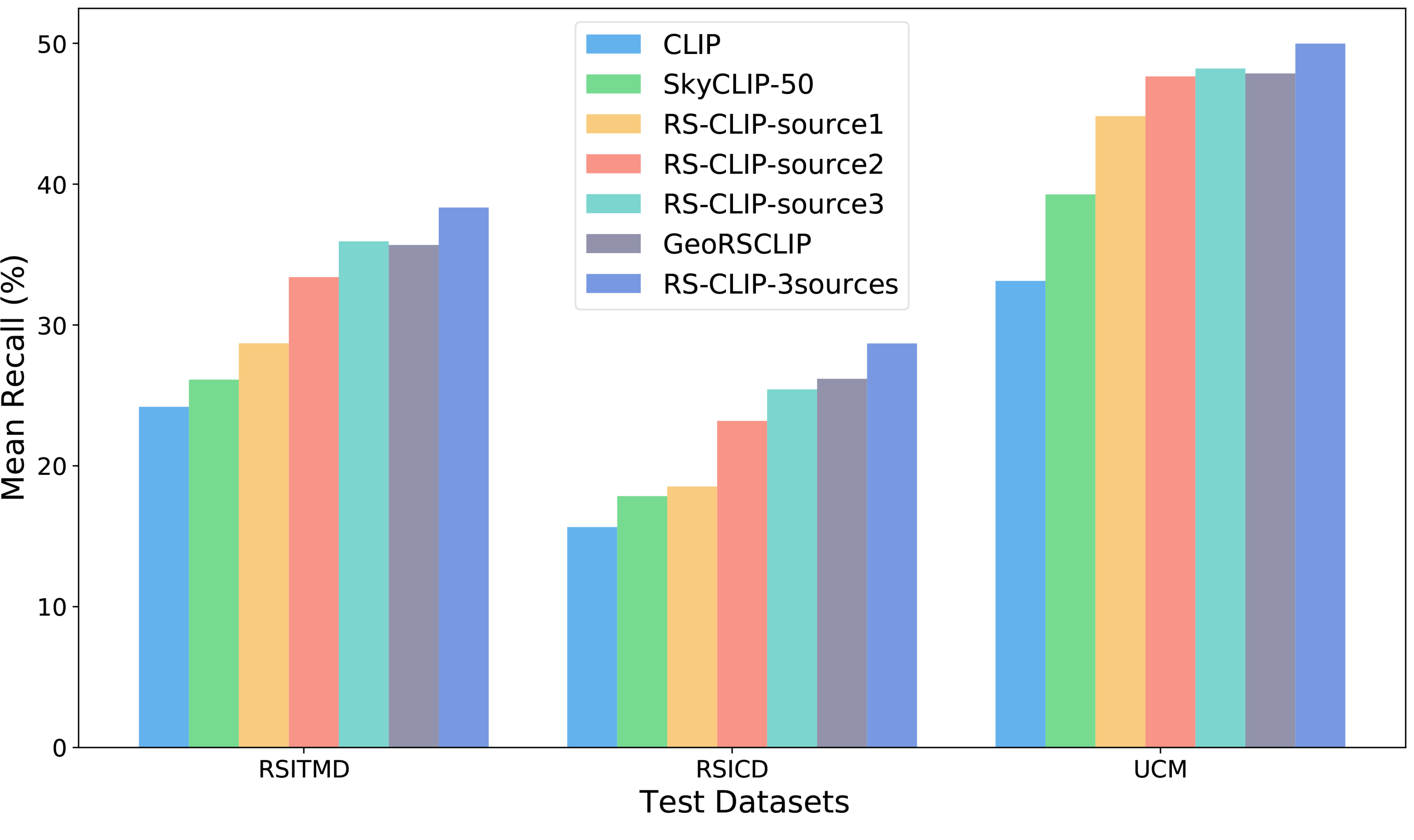}
  \caption{The RS Cross-modal Text-Image Retrieval result of different RS CLIPs fine-tuned with different training data. }
  \label{fig:mean_recall_comparison_Stage_1}
\end{figure}

\textbf{Validation of the Effectiveness of Different Descriptions.} To validate the effectiveness of our description strategies, we fine-tuned CLIP ViT-B-32 using data from three sources: source1 (R-M Relay), source2 (LLaVA-1.6 generation), and source3 (Kosmos-2 generation). We first trained CLIP on data from each source individually, resulting in RS-CLIP-source1, RS-CLIP-source2, and RS-CLIP-source3. Then, we trained CLIP on data from all three sources combined, resulting in RS-CLIP-3sources. Experimental results in Figure \ref{fig:mean_recall_comparison_Stage_1} showed that each source significantly improved CLIP's performance on the RS image-text retrieval task, while RS-CLIP-3sources outperformed all previous RS-CLIP models with less training data, demonstrating the complementary nature and effectiveness of these strategies.

\subsection{Stage 2: Caption Summarization}  

Although the captions from the first stage were sufficient to train the best RS CLIP, they had limitations: (1) Rule-based captions lacked sentence variety and semantic diversity; (2) MLLMs occasionally generated captions with hallucinations, such as city names; (3) Llava-1.6 generated captions averaging 150 words, exceeding the 77-token limit of CLIP's text encoder.

\begin{figure}[htb]
  \centering
  \includegraphics[width=0.5\textwidth]{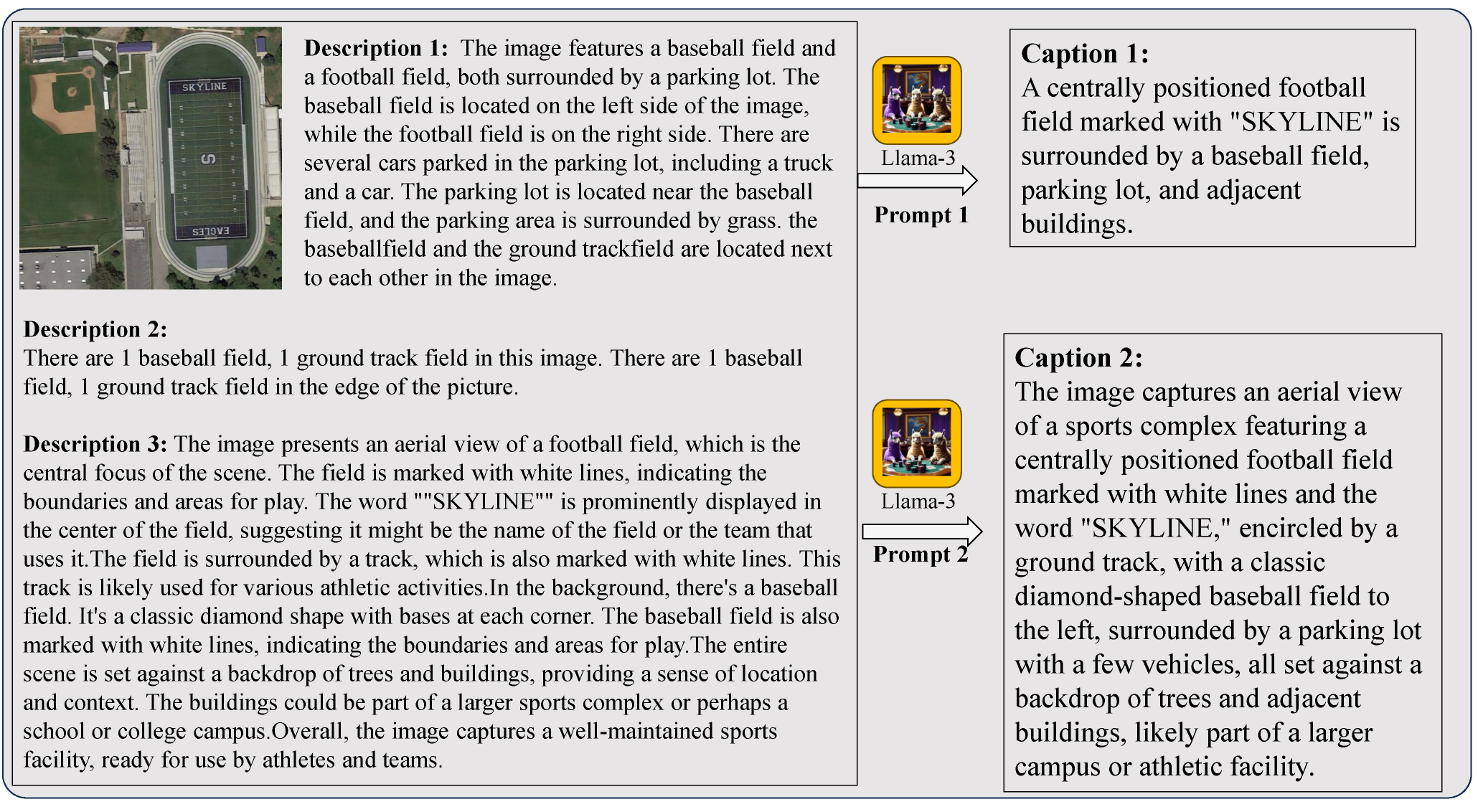}
  \caption{Pipeline of stage 2. Description 1 refers to the caption generated by Kosmos-2, Description 2 to the caption generated by the R-M Relay method, and Description 3 to the caption generated by Llava 1.6. Llama-3 generates different styles of comprehensive captions based on the complementary information from these three descriptions, guided by prompt 1 and prompt 2. The version of Llama-3 used here is Llama-3-8B-instruct.}
  \label{fig:pipeline_of_stage_2}
\end{figure}

To address these limitations, we use LLaMA-3-8B-Instruct to extract comprehensive visual information from the three captions and condense lengthy ones (see Figure \ref{fig:pipeline_of_stage_2}). This process consolidates complementary multi-perspective information from the initial stage into a unified description for each image. Carefully designed prompts filter out non-visual information, and long captions are compressed to shorter ones, ensuring compatibility with models like CLIP while preserving essential details.

When generating summarized captions, LLMs often produce a non-unimodal distribution~\citep{ic3}, with multiple peaks or modes. This makes single-prompt generation converge on a local maximum, resulting in "averaged" descriptions that fail to capture the diversity of the original text. To better reflect the complexity of the source text, strategies such as repeated sampling~\citep{ic3} can enhance diversity. 
To avoid randomness in repeated sampling, we use different prompts to generate summary captions, broadening the semantic space and better capturing textual complexity. Specifically, we generate two captions with distinct styles for each image using two prompts. For one caption, five possible options are generated, from which one is randomly selected for CLIP training. \textit{It is important to clarify that we sampled 1 out of every 5 captions for training the CLIP model to explore the effectiveness of dataset construction strategies. In contrast, the CoCa model was trained using the entire dataset. The final released dataset will also include all available data.}

It is widely recognized that prompts play a crucial role in the performance of LLMs. To maximize the potential of LLMs, we employed the CO-STAR Framework~\citep{co-star}, which secured the top results in the Prompt Engineering competition organized by the Government Technology Agency of Singapore (GovTech). The \textbf{CO-STAR} framework facilitates the construction of effective prompts by considering elements such as \textbf{C}ontext, \textbf{O}bjective, \textbf{S}tyle, \textbf{T}one, \textbf{A}udience, and \textbf{R}esponse. This structured approach enhances the quality and relevance of the responses generated by LLMs. After multiple experiments, we designed 2 prompts as follows.

\begin{algorithm}
\caption{Annotation to Instruction (A2I)}
\label{alg:a2i}
\SetAlgoNlRelativeSize{-1}
\SetKwFunction{FMain}{annotation\_to\_instruction}
\SetKwFunction{FExtract}{extract\_prompts}
\SetKwFunction{FGenerate}{generate\_instructions}
\SetKwProg{Fn}{Function}{:}{\KwRet{instructions}}

\Fn{\FMain{dataset, dataset\_type}}{
    prompts $\gets$ \FExtract{dataset, dataset\_type}\\
    instructions $\gets$ \FGenerate{prompts, dataset\_type}
}

\Fn{\FExtract{dataset, dataset\_type}}{
    prompts $\gets$ []\\
    \For{each image in dataset}{
        \If{dataset\_type == 'classification'}{
            prompts.append(image['category'])
        }
        \ElseIf{dataset\_type == 'object detection'}{
            prompts.append(f"obj[category] [location]")
        }
        \ElseIf{dataset\_type == 'segmentation'}{
            bbox\_list $\gets$ M2B(masks)
            prompts.append(f"obj[category] [location]")
        }
    }
    \Return prompts
}

\Fn{\FGenerate{prompts, dataset\_type}}{
    instructions $\gets$ []\\
    \For{each description in prompts}{
        \eIf{dataset\_type == 'classification'}{
            Instruction = "Describe image with [prompt] in detail:"
        }{
        \eIf{object\_number <= 2}{
            Instruction1 = "where is/are the [prompt]?"
            
            Instruction2 = "Describe image with [prompt] in detail:"
        }{
            Instruction = "Describe image with [prompt] in detail:"
        }
        }
    }
}

\end{algorithm}

\begin{itemize}[label={}]
    \item \textbf{Prompt 1}: {\slshape I am creating a RS image-text matching dataset. Each image has three different captions provided by different people, describing the same visual content from different perspectives. \textbf{Please summarize the main features of the following detailed description in a single sentence.} The sentence should use descriptive language with rich adjectives and prepositional phrases, include specific information about colors and spatial relationships, and maintain a clear and neutral tone. Use a professional and descriptive writing style, with a concise and content-rich tone. The target audience is researchers in the field of RS image analysis. Provide only the extracted sentence without any additional explanations or introductory text.}
\end{itemize}

\begin{itemize}[label={}]
    \item \textbf{Prompt 2}: {\slshape I am creating a RS image-text matching dataset. Each image has three different captions provided by different people, describing the same visual content from different perspectives. \textbf{Summarize the following three captions into one detailed sentence, including as many key visual elements as possible.} Use a professional and descriptive writing style. The tone should be concise and informative, suitable for researchers in the field of RS image analysis. Generate five detailed descriptions, each focusing on slightly different details to provide complementary information. Each description should be complete and complementary, ensuring full information coverage. \textbf{After generating the five detailed descriptions, randomly select one and output only that version}.
}
\end{itemize}

Finally, we generated captions with multiple styles for RS images. We fine-tuned CLIP using these data and tested their performance. The experimental results show that both prompts generated high-quality captions, significantly improving the performance of CLIP (see Table \ref{table:AblationOfStrategies}).

\begin{figure}[htbp]
  \centering
  \includegraphics[width=0.5\textwidth]{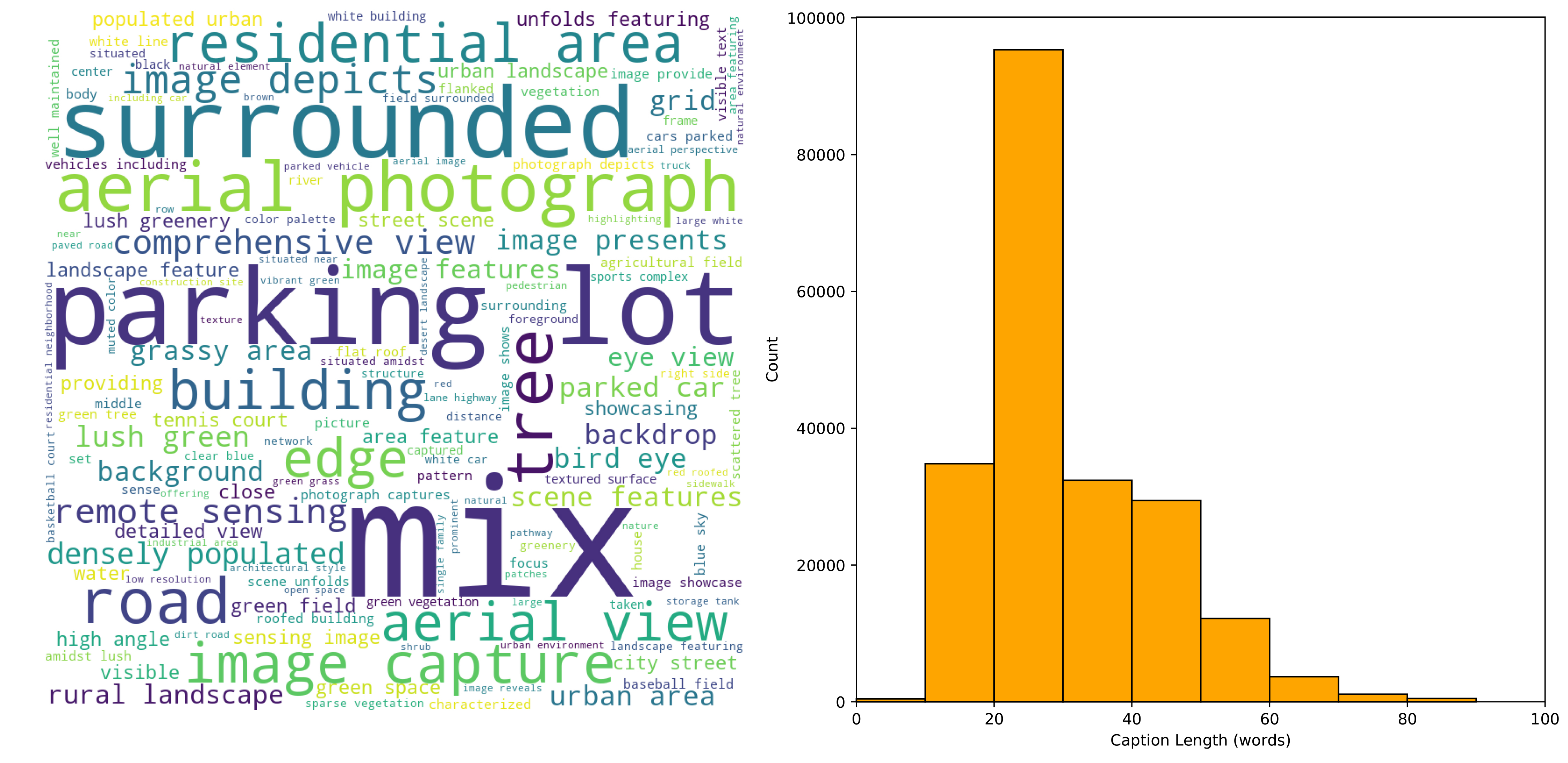}
  \caption{Wordcloud and Caption Length Distribution of HQRS-210K dataset. }
  \label{fig:WordCloud_and_CaptionLengthDistribution}
\end{figure}

\subsection{HQRS-IT-210K Dataset Description}
We finally got the HQRS-IT-210K dataset with 210556 images. For each image, we used two types of prompts to guide the LLM, generating 5 captions with one prompt and 1 caption with the other. This process resulted in a total of 6 captions per image, bringing the overall number of captions to approximately 1.26 million. As shown in Figure \ref{fig:CaptionCompairation}, the captions are naturally varied in structure and contain rich visual and semantic information. On average, each caption contains approximately 35.6 words, which is significantly longer than the 12-word average length of captions in human-annotated datasets.

As shown in Figure \ref{fig:WordCloud_and_CaptionLengthDistribution}, the word cloud of our captions demonstrates their diversity, descriptive precision, and contextual depth, highlighting the comprehensiveness and professionalism of our captioning approach. The caption length distribution indicates that most captions are rich in information and can be fully utilized by models like CLIP and CoCa.

\subsection{RS Long-Text image Retrieval Test Set}

\begin{figure}[tbp]
  \centering
  \includegraphics[width=0.5\textwidth]{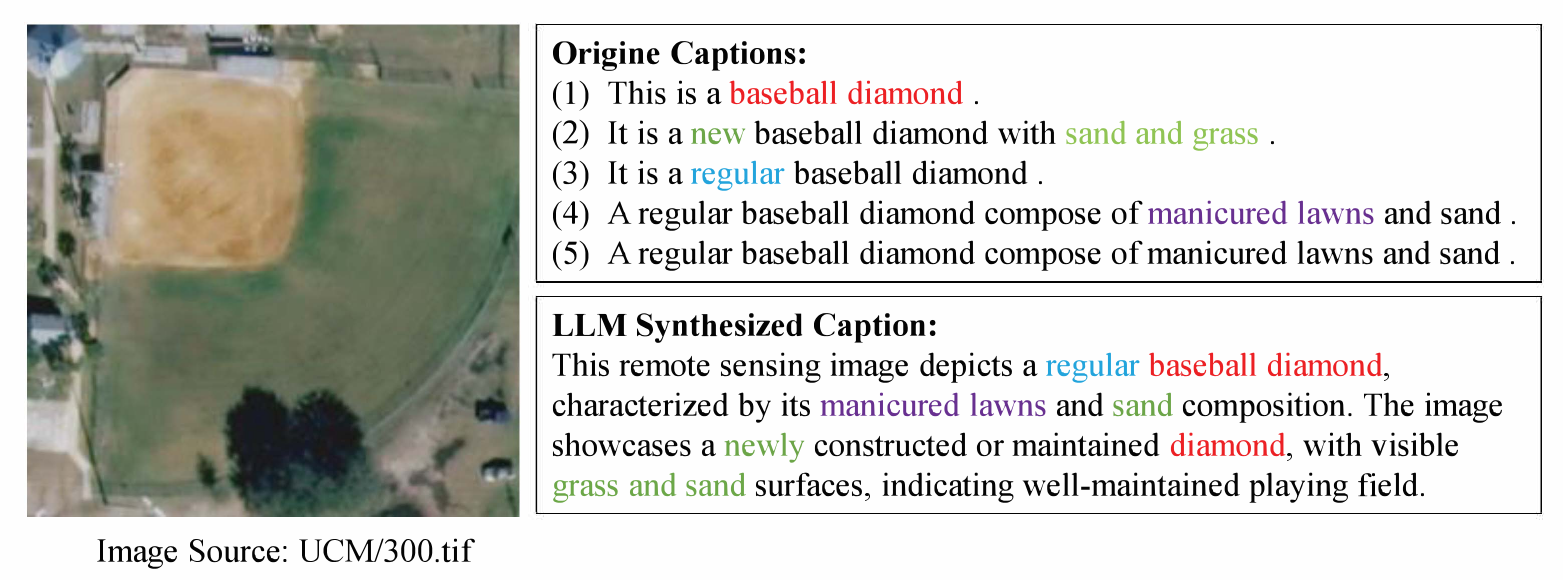}
  \caption{A Demo of Original Captions in UCM and Rewriting Results of Long Caption.}
  \label{fig:A Demo of LongRET3}
\end{figure}

While testing the retrieval performance of CLIP, we identified a significant limitation in existing RS image-text retrieval test datasets: the captions are too short (approximately 12 words). This leads to two issues: (1) short captions fail to fully describe images, causing category ambiguity, and (2) they are inadequate for evaluating CLIP's performance on long-text cross-modal retrieval tasks, as longer captions provide more complex and complete information. To address this, we constructed an RS long-text image retrieval test set.

A simple approach is to concatenate the five short captions into a single long-text caption. However, this has major drawbacks: (1) concatenation introduces redundancy and repetition, reducing overall information density. For example, in Figure \ref{fig:A Demo of LongRET3}, Sentence (1) and Sentence (3) are semantically similar, while Sentence (4) and Sentence (5) are identical. Such repetition dilutes key information and decreases the proportion of effective content. (2) Concatenated captions often lack coherence and fluency, as each short caption is originally an independent description, leading to logical disconnections when linked together.

\begin{figure}[htb]
  \centering
  \includegraphics[width=0.5\textwidth]{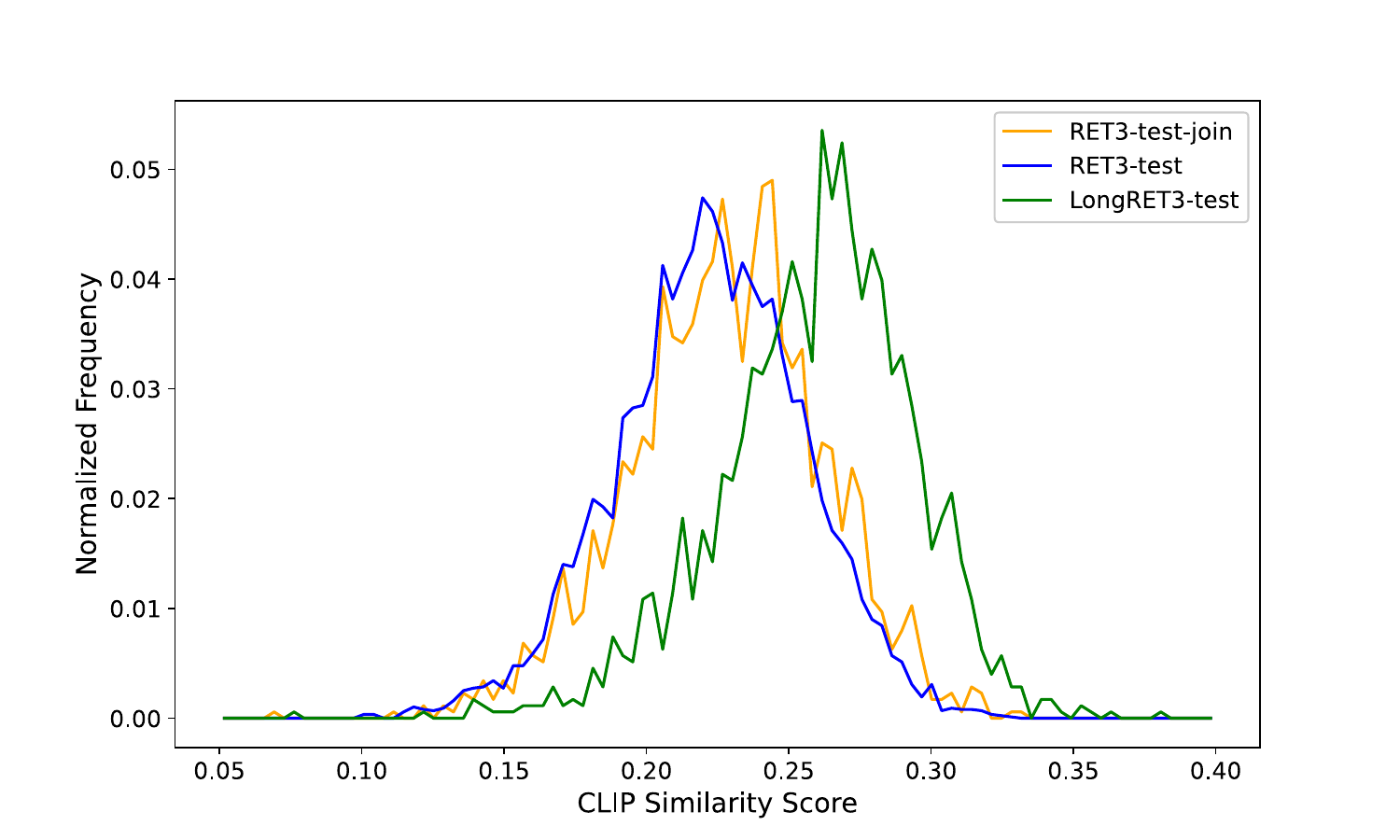}
  \caption{CLIP Similarity Score Distribution of \textbf{RET3-test}, \textbf{RET3-test-join}, and \textbf{LongRET3-test}.}
  \label{fig:SimilarityScoreDistribution}
\end{figure}

To address these shortcomings, LLMs can be employed for caption integration and rewriting~\citep{ic3}, effectively resolving redundancy, coherence, and error issues (see Figure \ref{fig:A Demo of LongRET3}). 
Firstly, LLMs enhance information density by eliminating redundancy and repetition through semantic understanding and information extraction, integrating key details from the original captions into a concise, informative long caption. Secondly, LLMs improve coherence and fluency by connecting independent short sentences with appropriate transitions, resulting in semantically coherent and logically consistent long captions. 
Thirdly, LLMs correct grammatical and semantic errors in the original captions, which are common due to non-native annotators. For example, in Figure \ref{fig:A Demo of LongRET3}, Sentence (4) and Sentence (5) contain grammatical errors, which are resolved after LLM rewriting.

We utilized LLaMA-3-8B-Instruct to combine and rewrite the five captions for each image in the test sets of three widely used benchmark datasets (RSITMD, RSICD, and UCM) into a single long-text caption. This approach has key advantages: (1) these datasets are well-established and widely recognized; (2) most images include five complementary captions, and combining them with LLM creates comprehensive, high-quality long-text captions at a low cost.

To ensure that the LLM strictly adheres to the content of the original captions, we carefully designed the following prompt.

\begin{itemize}[label={}]
    \item \textbf{Prompt}: {\slshape I am creating a RS image-long text matching dataset based on an existing RS image-short text matching dataset. In the short text matching dataset, each image has 5 different short captions provided by different individuals, describing the same visual content from different perspectives. Please summarize the following 5 short captions into one detailed long caption, ensuring that the information from each short caption is included in the long caption. \textbf{The summarization process must strictly adhere to the information given in the 5 short captions without any additional extrapolation.} Use a professional and descriptive writing style, with a concise and content-rich tone. The target audience is researchers in the field of RS image analysis. Do not include any additional explanations or introductory text in your response.} 
\end{itemize}

As demonstrated in Figure \ref{fig:A Demo of LongRET3}, the advantages of this approach are evident. The LLM-rewrite long caption effectively integrates key information from each short caption, enhancing the overall information density. This comprehensive description provides a more complete image representation, reducing potential category ambiguities. Additionally, by naturally combining sentences, the LLM addresses the coherence and fluency issues that can arise from simple concatenation, ensuring a more seamless and readable caption. Moreover, the potential grammatical and semantic errors in the original captions can also be resolved.

For simplification, we collectively refer to the test datasets from three benchmark datasets as \textbf{RET3-test}. The data where the five captions corresponding to each image are concatenated into a "pseudo-long caption" is termed \textbf{RET3-test-join}. The dataset with long captions rewritten by the LLM is referred to as \textbf{LongRET3-test}. We utilized CLIP ViT-L-14 to calculate the image-text similarity scores for RET3-test, RET3-test-join and LongRET3-test. These scores were used to evaluate their quality. To minimize the impact of hallucinations generated by large models, we manually reviewed each caption to ensure that no obvious issues were present.

As illustrated in Figure \ref{fig:SimilarityScoreDistribution}, the distribution of image-text similarity scores for the original RET3-test closely mirrors that of the RET3-test-join. The LongRET3-test not only maintains a similar distribution to the first two sets but also demonstrates an overall increase in similarity scores. The primary reason for this improvement is that combining the information from the five distinct captions reduces category ambiguity, resulting in captions that are more accurately aligned with the image content.

\section{EXPERIMENTS} 

To evaluate the effectiveness of our dataset on both discriminative and generative tasks, we fine-tuned the CLIP and CoCa models separately. For the CLIP model, we used only 1/6 of the entire dataset, with one caption per image, to highlight the high quality of our dataset when aligning with similar studies. In contrast, for the CoCa model, due to the higher data demands of generative tasks, we utilized the full dataset of approximately 1.26 million captions for training. Since similar studies primarily focus on the CLIP model, this paper also emphasizes the study and comparison of the CLIP model.

\subsection{Caption Fusion Strategy for CLIP Training Data}

\begin{figure}[htb]
  \centering
  \includegraphics[width=0.5\textwidth]{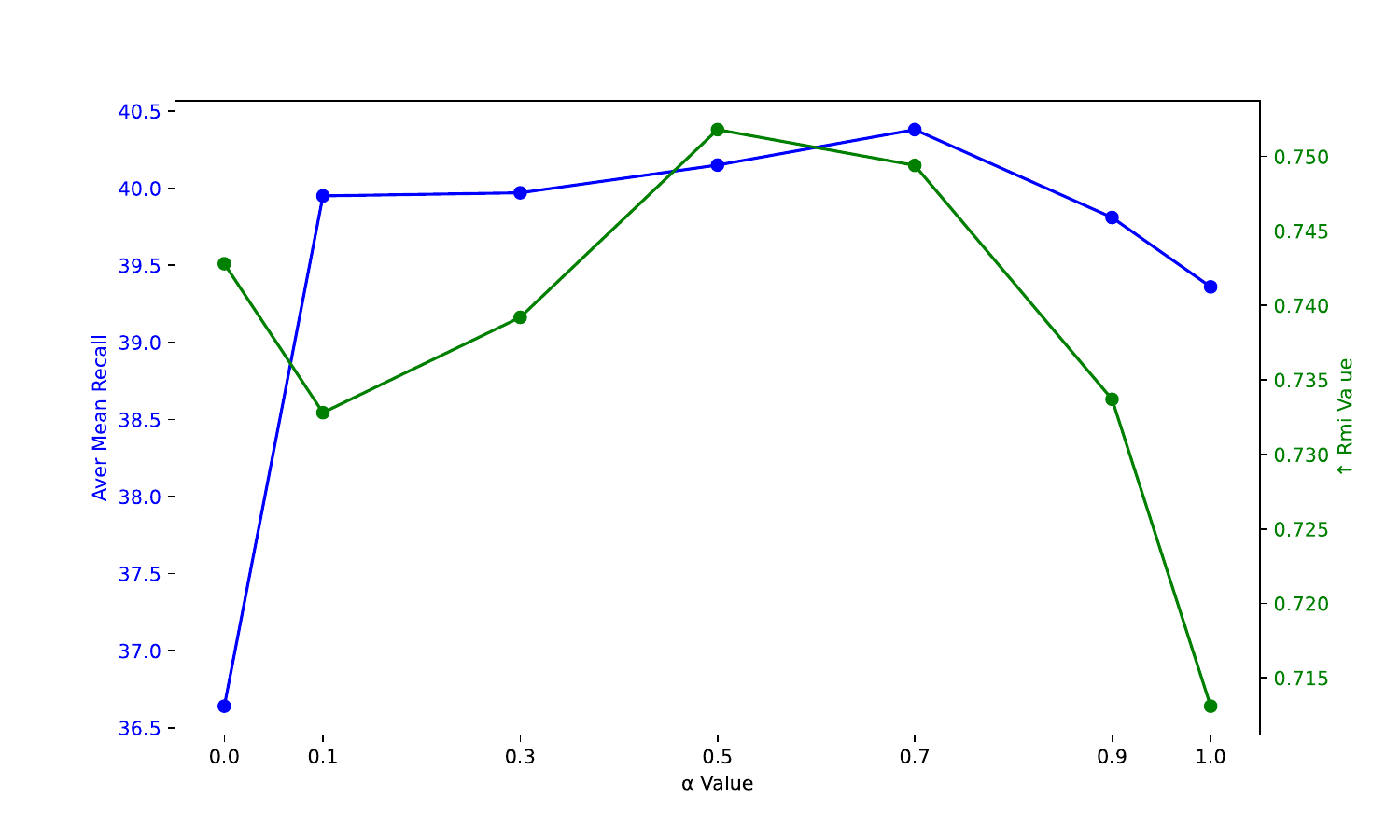}
  \caption{Effectiveness of Different \(\alpha\) Value.  }
  \label{fig:ablation_of_α}
\end{figure}

When selecting the training data for CLIP, we employed a very simple yet highly effective data fusion strategy. In the second stage, we used two prompts to guide the LLM in generating distinct caption styles, simulating real-world diversity in descriptions. For Prompt 1, the LLM generated five captions, from which one was randomly selected. Now, each image is associated with two captions of different styles generated from two distinct prompts. To further refine this, we tried to fuse these two captions into a single caption, ensuring that each image is ultimately associated with one unified caption.
Through several attempts, we found that a simple fusion strategy—\textit{randomly selecting one caption from two different styles for each image with a probability of 
$\alpha$ and using the selected one as the sole caption}—is effective.
$\alpha$ is the proportion of “prompt 2 generated captions” in the final caption data, where $\alpha$ = 0 means only “prompt 1 generated captions” were used.
Experimental results show that this very simple fusion method significantly improves model performance without increasing training costs and outperforms models trained on any single caption style(See Figure \ref{fig:ablation_of_α}).

The effectiveness of this strategy is easy to understand. This approach allowed the LLM to focus on different aspects of the image, resulting in varied captions in content, style, and emphasis. When the training data includes multiple caption styles, it can enhance model performance from two perspectives. First, data diversity allows the model to capture a wider range of information, leading to better overall performance. Second, the variety in captions acts as a regularization mechanism, preventing the model from overfitting to a single style and improving its generalization ability.

\subsection{Implementation Details of Experiments}

\textit{CLIP training.} In RemoteCLIP, SkyCLIP and GeoRSCLIP, models were trained using the full fine-tuning approach. To ensure a fair comparison, we also used this method. The CLIP ViT-B-32 and CLIP ViT-L-14 models were fully fine-tuned on our dataset. 
To emphasize the intrinsic effectiveness of our dataset, we did not use any data augmentation techniques or perform any specific hyperparameter tuning during training. We randomly selected 15\% of the data as the validation set, with the remaining data used for training. The training process utilized a cosine learning rate scheduler, mixed precision (AMP) mode, and the AdamW optimizer~\citep{AdamW}. The modulal interaction was conducted using the InfoNCE loss~\citep{InfoLoss}. For ViT-B-32, the learning rate was set to 2e-5, and the batch size to 256. For ViT-L-14, the learning rate was set to 1e-6, and the batch size to 32. Ultimately, we obtained HQRS-CLIP (ViT-B-32) and HQRS-CLIP (ViT-L-14). \textbf{Compared to RemoteCLIP's 233.4 hours of training time, our HQRS-CLIP (ViT-L-14) training only required approximately 1.5 hours.}

\textbf{CoCa training.} The CoCa model was also trained using full fine-tuning . We utilized all 1.26 million image-text pairs without applying any data augmentation strategies. The AdamW optimizer was also used during training, with the CoCa-contrastive-loss weight set to 1 and the CoCa-caption-loss weight set to 2. We only trained the CoCa ViT-L-14 model with a learning rate of 2e-5 and a batch size of 32. During the benchmark testing, we utilized the training data from the corresponding benchmark datasets and adjusted the loss weights by setting the CoCa-contrastive-loss weight to 0 and the CoCa-caption-loss weight to 1. This adjustment was made to enhance the model's caption generation capability specific to the respective datasets.

Both the CLIP and CoCa models were trained on a single RTX 4090 24 GB GPU, and their training code were all based on OpenCLIP\footnote{\url{https://github.com/mlfoundations/open_clip}}.

We conducted model testing on the following tasks. 

\textbf{HQRS-CLIP: RS Cross-modal Text–Image Retrieval (RSCTIR)}: This task encompasses both text-to-image and image-to-text retrieval using the UCM, RSICD, and RSITMD datasets. We also assessed zero-shot retrieval to test generalization to unseen datasets. Performance was measured using R@1 for both retrieval directions and mean recall.

\textbf{HQRS-CLIP: RS Cross-modal Long text-Image Retrieval}:  To better evaluate the model's ability to understand longer and more comprehensive descriptions, we created the first remote sensing (RS) long-text image retrieval test set by merging and rewriting existing datasets. The new captions average 72 words, significantly longer than the typical 12-word captions, and offer improved alignment with images. We evaluated performance with i2t-R@1, i2t-R@5, i2t-R@10, t2i-R@1, t2i-R@5, t2i-R@10, and mean recall metrics.

\textbf{HQRS-CLIP: Semantic Localization (SeLo)}: SeLo tasks are defined as using cross-modal information to locate semantically similar regions in large-scale RS scenes, considered a more advanced retrieval task than RSCTIR. AIR-SLT is currently the only large-scale multimodal semantic localization test set. We employed four metrics: Rsu (Significant Area Proportion), Ras (Attention Shift Distance), Rda (Discrete Attention Distance), and Rmi (a composite metric) to comprehensively assess performance.

\textbf{HQRS-CLIP: Zero-shot Classification (ZSC)}: Following the extended definition of zero-shot learning, we tested the model's ability to generalize to unseen datasets using AID, RESISC45, and EuroSAT, evaluating with top-1 accuracy. These three datasets were also tested by models such as SkyCLIP, GeoRSCLIP, and RemoteCLIP. We selected them for testing to ensure a fair comparison with these models.

\textbf{HQRS-CLIP: Few-shot Classification (FSC)}: To assess the model's capability with limited labeled data, we conducted 5-way 1-shot and 5-way 5-shot classification on RESISC45 and UCM datasets, based on the benchmark introduced by \cite{Few-shot-Classi}. This evaluated the feature extraction performance of RS CLIP’s image encoder under few-shot conditions.

\textbf{RS-CoCa: RS Image Captioning (RSIC)}: For the RS-CoCa model, we evaluated its performance on RS image captioning task. We used the widely adopted RSICD and UCM-caption datasets as benchmarks to facilitate comparison with other studies. Similarly, we employed commonly used evaluation metrics, including BLEU, METEOR, ROUGE-L, CIDEr, and SPICE. Additionally, to provide a more intuitive demonstration of our model's strong performance, we conducted qualitative analysis by showcasing examples of captions generated by RS-CoCa.

\subsection{Experimental Results}

For RS-CoCa, we primarily evaluated its performance on the RS image captioning task to facilitate comparisons with similar VLFMs. The image backbone of RS-CoCa is ViT-L-14. For HQRS-CLIP, We evaluated its performance on a series of vision-language tasks. The image backbone of HQRS-CLIP is ViT-B-32 unless we specify. All the test sets have been used in previous related RS CLIP studies, and we utilize the same test sets to ensure a fair peer comparison. \textit{Experimental results indicate that HQRS-CLIP shows superior capabilities while using only 4.2\% of the data of GeoRSCLIP, 16\% of SkyCLIP-50's data and 25\% of RemoteCLIP's data.}

\begin{table} 
\centering
\caption{Results of Zero-shot RSCTIR Task. The best result is in \textbf{bold}. \textbf{Training Pairs} indicates the number of training image-text pairs used for training the RS CLIP models.}
\fontsize{7}{8}\selectfont
\begin{tabular}{lccccc}
\toprule
\textbf{\makecell{Test \\ Dataset}} & \textbf{Models} & \textbf{\makecell{Training \\ Pairs}} & \makecell{\textbf{I2T} \\ \textbf{R@1}} & \makecell{\textbf{T2I} \\ \textbf{R@1}} & \textbf{Mean Recall} \\
\midrule

\multirow{6}{*}{RSITMD} & CLIP & - & 9.51 & 8.81 & 24.19 \\
 & SkyCLIP-50 & 1.2 Million+ & 8.63 & 11.02 & 26.12 \\
 & GeoRSCLIP & 5 Million+ & 19.03 & 14.16 & 36.68 \\
 & Ours & \textbf{210K+} & \textbf{20.58} & \textbf{17.30} & \textbf{40.15} \\
 \cmidrule(lr){2-6}
 & Ours (ViT-L-14) & \textbf{210K+} & \textbf{21.90} & \textbf{19.16} & \textbf{40.63} \\
\cmidrule(lr){1-6}

\multirow{6}{*}{RSICD} & CLIP & - & 5.31 & 5.78 & 15.74 \\
 & SkyCLIP-50 & 1.2 Million+ & 5.86 & 6.31 & 17.85 \\
 & GeoRSCLIP & 5 Million+ & 11.53 & 9.52 & 26.18 \\
 & Ours & \textbf{210K+} & \textbf{15.55} & \textbf{11.20} & \textbf{29.37} \\
 \cmidrule(lr){2-6}
 & Ours (ViT-L-14) & \textbf{210K+} & \textbf{15.55} & \textbf{11.95} & \textbf{30.33} \\
\cmidrule(lr){1-6}

\multirow{6}{*}{UCM} & CLIP & - & 9.52 & 8.67 & 33.13 \\
 & SkyCLIP-50 & 1.2 Million+ & 10.00 & 9.90 & 39.27 \\
 & GeoRSCLIP & 5 Million+ & 18.57 & 13.81 & 47.76 \\
 & Ours & \textbf{210K+} & \textbf{22.86} & \textbf{16.19} & \textbf{51.59} \\
 \cmidrule(lr){2-6}
 & Ours (ViT-L-14) & \textbf{210K+} & 22.38 & \textbf{16.67} & \textbf{53.33} \\
\bottomrule

\label{tab:Test Results of Zero-shot RSCTIR Task}
\end{tabular}
\end{table}

\subsubsection{RS Cross-modal Text-Image Retrieval}
\textbf{Zero-shot RSCTIR task}. Table \ref{tab:Test Results of Zero-shot RSCTIR Task} shows that HQRS-CLIP surpasses existing RS CLIP models on 3 benchmarks using zero-shot RSCTIR. Specifically, HQRS-CLIP achieves an average mean recall that is 16.02\% higher than the CLIP baseline, 12.62\% higher than SkyCLIP-50, and 3.50\% higher than GeoRSCLIP. Additionally, scaling up to CLIP ViT-L-14 further improves performance, increasing the average mean recall by 1.06\% compared to HQRS-CLIP ViT-B-32, demonstrating the benefits of larger models in leveraging our dataset.

\begin{table}
\centering
\footnotesize
\caption{Results of Fine-tuned RSCTIR tasks. The best result is in \textbf{bold}. \textbf{Training Pairs} indicates the number of training image-text pairs used for training the RS CLIP models.}
\label{tab:Results of RSCTIR task}
\fontsize{7}{8}\selectfont
\begin{tabular}{lcccccc}
\toprule
\textbf{\makecell{Image \\ Backbone}} & \textbf{\makecell{Test \\ Dataset}} & \textbf{Models} & \textbf{\makecell{Training \\ Pairs}} & \makecell{\textbf{I2T} \\ \textbf{R@1}} & \makecell{\textbf{T2I} \\ \textbf{R@1}} & \makecell{\textbf{Mean} \\ \textbf{Recall}} \\
\midrule
\multirow{11}{*}{ViT-B-32} & \multirow{4}{*}{RSITMD} & RemoteCLIP & 820 K & 27.88 & 22.17 & 49.38 \\
 &  & GeoRSCLIP & 5 Million+ & 30.09 & 23.54 & 50.10 \\
 &  & Ours & \textbf{210K+} & \textbf{35.18} & \textbf{29.65} & \textbf{54.53} \\
\cmidrule(lr){2-7}
 & \multirow{4}{*}{RSICD} & RemoteCLIP & 820 K+ & 17.02 & 13.71 & 35.26 \\
 &  & GeoRSCLIP & 5 Million+ & \textbf{22.14} & 15.26 & 38.00 \\
  &  & Ours & \textbf{210K+} &  20.86 & \textbf{15.61} & \textbf{38.08} \\
\cmidrule(lr){2-7}
 & \multirow{3}{*}{UCM} & RemoteCLIP & 820 K+ & 20.48 & 18.67 & 56.36 \\
  &  & Ours & \textbf{210K+} & \textbf{22.86} & \textbf{19.14} & \textbf{57.21} \\
\midrule
\multirow{9}{*}{ViT-L-14} & \multirow{3}{*}{RSITMD} & RemoteCLIP & 820 K+ & 28.76 & 23.76 & 50.52 \\
  &  & Ours & \textbf{210K+} & \textbf{34.07} & \textbf{31.02} & \textbf{57.05} \\
\cmidrule(lr){2-7}
 & \multirow{3}{*}{RSICD} & RemoteCLIP & 820 K & 18.39 & 14.73 & 36.35 \\
 &  & Ours & \textbf{210K+} & 21.5 & \textbf{18.54} & \textbf{40.19} \\
\cmidrule(lr){2-7}
 & \multirow{3}{*}{UCM} & RemoteCLIP & 820 K & 19.05 & 17.71 & 54.68 \\
 &  & Ours & \textbf{210K+} & 20.48 & \textbf{19.52} & \textbf{58.32} \\
\bottomrule
\end{tabular}
\end{table}

\textbf{Fine-tuned RSCTIR task}. For a fair comparison with RemoteCLIP and GeoRSCLIP, we fine-tuned HQRS-CLIP on the RET-3 \footnote{RET-3 provided by RemoteCLIP includes 68,565 deduplicated image-text pairs from RSITMD, RSICD, and UCM.}. As shown in Table \ref{tab:Results of RSCTIR task}, HQRS-CLIP achieved the best results across the majority of metrics. Notably, HQRS-CLIP excels in the Text-to-Image retrieval task. Overall, HQRS-CLIP outperforms RemoteCLIP by 2.94\% and GeoRSCLIP by 2.26\% in average mean recall. Furthermore, HQRS-CLIP (ViT-L-14) exceeds RemoteCLIP (ViT-L-14) by 4.67\%, establishing it as the state-of-the-art model on all three benchmarks.

\subsubsection{RS Cross-modal Long-Text Image Retrieval}

\textit{\textbf{Before reporting these experimental results, it is important to note that these results are intended for preliminary evaluation only and might not be fully conclusive, as HQRS-CLIP were trained on longer texts. The introduction of this test benchmark and the initial evaluation aim to supplement the current text image retrieval test data and provide insights for the RS community.}}

\begin{table*}[hbt]
\centering
\footnotesize
\caption{Results of LongRET3-test Retrieval. The best result is in \textbf{bold}. \textbf{FT} means that the model was fine-tuned on the RET-3 training data.}
\footnotesize
\begin{tabular*}{\textwidth}{@{\extracolsep{\fill}}lp{1.0cm}p{1.0cm}p{1.0cm}p{1.0cm}p{1.0cm}p{1.0cm}p{1.2cm}}
\toprule
Model & I2T-R@1 & I2T-R@5 & I2T-R@10 & T2I-R@1 & T2I-R@5 & T2I-R@10 & Mean-Recall \\
\midrule
CLIP-Baseline   & 4.79  & 16.01  & 24.44  & 3.25  & 14.13  & 22.91  & 14.25 \\
SkyCLIP-50       & 5.01  & 16.13  & 27.12  & 5.41  & 16.92  & 28.21  & 16.47 \\
GeoRSCLIP        & 11.40 & 30.48  & 43.82  & 9.86  & 27.12  & 38.92  & 26.93 \\
Ours             & \textbf{14.70} & \textbf{34.93}  & \textbf{48.43}  & \textbf{15.84} & \textbf{37.66}  & \textbf{51.68}  & \textbf{33.87} \\
\midrule
RemoteCLIP(FT)       & 12.71 & 35.44  & 52.02  & 9.86  & 30.54  & 45.41  & 31.00 \\
Ours(FT)   & \textbf{16.58} & \textbf{44.33}  & \textbf{60.57}  & \textbf{16.01} & \textbf{40.85}  & \textbf{55.84}  & \textbf{39.03} \\
\bottomrule
\label{tab:Test Results of LongRET3-test}
\end{tabular*}
\end{table*}

We evaluated the existing RS CLIP models on the LongRET3-test dataset, with the results presented in Table \ref{tab:Test Results of LongRET3-test}. Among the models not fine-tuned on RET-3, HQRS-CLIP outperformed SkyCLIP-50 by 17.4\% and GeoRSCLIP by 6.94\% in mean recall. For models fine-tuned on RET-3, HQRS-CLIP achieved a mean recall that is 8.03\% higher than RemoteCLIP. 
Moreover, whether fine-tuned on RET-3 or not, HQRS-CLIP consistently achieved the best recall scores across all metrics, demonstrating its superior performance in long-text image-text retrieval.

\subsubsection{Semantic Localization Task}

Table \ref{tab:result_SeLo} demonstrates that HQRS-CLIP outperforms all existing models on the AIR-SLT dataset across all four metrics. HQRS-CLIP achieves the highest Rsu, indicating maximum attention to the ground truth (GT) region, the lowest Ras, reflecting minimal attention shift distance, and the best Rda, showing highly concentrated attention within the GT region. Additionally, HQRS-CLIP leads in the Rmi score, highlighting its exceptional performance in semantic localization.

\begin{table}[hbt]
\centering
\caption{Results of SeLo task. The best result is in \textbf{bold}.} 
\label{tab:result_SeLo} 
\footnotesize 
\begin{tabular}{lccccc}
\toprule
\textbf{Method} & \textbf{Training pairs} & \textbf{Rsu $\uparrow$} & \textbf{Ras $\downarrow$} & \textbf{Rda $\uparrow$} & \textbf{Rmi $\uparrow$} \\
\midrule
CLIP-Baseline & - & 0.7188 & 0.3006 & 0.6992 & 0.7071 \\
SkyCLIP-50    & 1.3 Million+    & 0.7341 & 0.2997 & 0.6948 & 0.7125 \\
RemoteCLIP & 820K+ & 0.7365 & 0.3008 & 0.6928 & 0.7125 \\
GeoRSCLIP & 5 Million+ & 0.7546 & 0.2610 & 0.7180 & 0.7400 \\
ours & \textbf{210K+}      &\textbf{ 0.7635} & \textbf{0.2488} & \textbf{0.734}  & \textbf{0.7518} \\
\bottomrule
\end{tabular}
\end{table}

\subsubsection{Zero-shot Classification}

Since our full dataset includes portions of these three datasets, we excluded overlapping data from our full dataset before independently training CLIP for testing to prevent potential data leakage and to ensure an unbiased comparison.

\begin{table}[htbp]
\centering
\caption{Results of Zero-shot Classification Tasks. The best result is in \textbf{bold}.}
\footnotesize
\begin{tabular}{lccccc}
\toprule
Method         & \makecell{Training \\ Pairs}   & AID   & RESISC45 & EuroSAT & Aver \\
\midrule
CLIP-Baseline  & -            & 66.22 & 60.90    & 47.21   & 58.11   \\
SkyCLIP-50     & 1.3 million  & 69.49 & 62.92    & 49.62   & 60.68   \\
RemoteCLIP     & 820K      & \textbf{92.87} & 70.84    & 35.01   & 66.24   \\
GeoRSCLIP      & 5 Million+   & 73.72 & 71.89    & \textbf{61.49}   & 69.03   \\
Ours           & 210 K        & 73.86 & \textbf{78.41}    & 60.55   & \textbf{70.94}   \\
\bottomrule
\end{tabular}
\label{tab:zero_shot_classification}
\end{table}

As shown in Table \ref{tab:zero_shot_classification}, HQRS-CLIP achieved the highest average zero-shot top-1 accuracy across the three datasets, outperforming GeoRSCLIP by 1.91\%, RemoteCLIP by 4.70\%, and SkyCLIP-50 by 10.26\%. Additionally, RemoteCLIP displayed inconsistent performance, achieving 92.87\% on the AID dataset but performing poorly on EuroSAT, likely due to training data imbalance. In contrast, HQRS-CLIP and GeoRSCLIP demonstrated similar robustness across all three datasets.

\subsubsection{Few-shot Classification}
Following \cite{Few-shot-Classi}, we conducted two types of few-shot classification tests. The first type involved testing on a data subset, the entire dataset, and all available classes. The second type comprised four settings: Setting A tested only the test subset without using the training data; Setting B evaluated all classes within the test subset, consisting of 10 and 6 classes for the two datasets, respectively; Setting C used the entire dataset with randomly selected 5 classes for each task; and Setting D tested the entire dataset with all classes distinguished, featuring 45 and 21 classes for the two datasets, respectively. Overall, Setting A aligns with standard practice, while Setting D represents a more extreme and challenging scenario involving larger datasets and more classes.

\begin{table*}[htbp]
\centering
\caption{Results of N-way K-shot Classification Accuracy. The best result is in \textbf{bold}.}
\label{tab:comparison}
\small
\begin{tabular}{p{1.8cm} p{1.6cm} p{0.3cm} p{1.6cm} p{1.6cm} p{0.3cm} p{1.6cm} p{1.6cm}}
\toprule
test set & method & way & \multicolumn{2}{c}{RESISC45} & way & \multicolumn{2}{c}{UCM} \\
\cmidrule(lr){4-5} \cmidrule(lr){7-8}
 &  &  & 1-shot & 5-shot &  & 1-shot & 5-shot \\
\midrule
test split & \textit{baseline} & 5 & 86.55 ± 0.18 & 91.06 ± 0.11 & 5 & 70.73 ± 0.16 & 82.92 ± 0.12 \\
\midrule
 & RemoteCLIP & 5 & 87.11$\pm$0.08 & 95.18$\pm$0.03 & 5 & - & - \\ 
 test split & GeoRSCLIP & 5 & 91.57$\pm$0.05 & \textbf{97.08$\pm$0.02} & 5 & 91.95$\pm$0.06 & 98.69$\pm$0.01 \\ 
 (Setting A)& SkyCLIP & 5 & 84.37$\pm$0.08 & 94.10$\pm$0.03 & 5 & 93.39$\pm$0.04 & 98.83$\pm$0.01 \\ 
 & Ours & 5 & \textbf{92.15$\pm$0.05} & 96.58$\pm$0.03 & 5 & \textbf{96.15$\pm$0.03} & \textbf{99.54$\pm$0.00} \\ 
\midrule
test split & RemoteCLIP & 10 & 81.44$\pm$0.04 & 91.91$\pm$0.01 & 6 & - & - \\ 
(all classes) & GeoRSCLIP & 10 & 82.32$\pm$0.05 & 93.25$\pm$0.01 & 6 & 92.23$\pm$0.04 & 98.29$\pm$0.01 \\ 
(Setting B) & SkyCLIP & 10 & 75.45$\pm$0.05 & 89.82$\pm$0.01 & 6 & 90.99$\pm$0.05 & 98.41$\pm$0.01 \\ 
 & Ours & 10 & \textbf{85.09$\pm$0.03} & \textbf{94.14$\pm$0.01} & 6 & \textbf{95.48$\pm$0.02} & \textbf{99.38$\pm$0.00} \\ 
\midrule
whole dataset & RemoteCLIP & 5 & 87.01$\pm$0.09 & 96.01$\pm$0.04 & 5 & - & - \\ 
(all classes) & GeoRSCLIP & 5 & 91.10$\pm$0.06 & 97.35$\pm$0.02 & 5 & 92.27$\pm$0.06 & 96.43$\pm$0.04 \\ 
(Setting C) & SkyCLIP & 5 & 86.05$\pm$0.08 & 95.74$\pm$0.03 & 5 & 91.67$\pm$0.06 & 97.01$\pm$0.03 \\ 
 & Ours & 5 & \textbf{93.05$\pm$0.05} & \textbf{97.82$\pm$0.02} & 5 & \textbf{95.23$\pm$0.05} & \textbf{97.64$\pm$0.04} \\ 
\midrule
whole dataset & RemoteCLIP & 45 & 62.43$\pm$0.03 & 81.18$\pm$0.01 & 21 & - & - \\ 
(all classes) & GeoRSCLIP & 45 & 66.24$\pm$0.03 & 84.24$\pm$0.01 & 21 & 77.05$\pm$0.03 & 91.04$\pm$0.01 \\ 
(Setting D) & SkyCLIP & 45 & 57.12$\pm$0.03 & 78.42$\pm$0.01 & 21 & 72.61$\pm$0.04 & 89.26$\pm$0.01 \\ 
 & Ours & 45 & \textbf{68.71$\pm$0.03} & \textbf{85.83$\pm$0.01} & 21 & \textbf{82.42$\pm$0.02} & \textbf{93.06$\pm$0.01} \\ 
\bottomrule
\end{tabular}
\end{table*}

Since our full dataset includes portions of the RESISC45 data, we first excluded all RESISC45 entries before fine-tuning CLIP to prevent data leakage and ensure an unbiased comparison. The experimental results demonstrate that HQRS-CLIP outperformed previous RS CLIP models in nearly all settings (15/16). Notably, in the 1-shot tasks, HQRS-CLIP significantly surpassed all existing RS CLIP models. These findings indicate that our image encoder has developed strong feature recognition and generalization capabilities on our dataset, indirectly validating the high quality of our dataset.

\subsubsection{RS image captioning}

\begin{figure}[htb]
  \centering
  \includegraphics[width=0.5\textwidth]{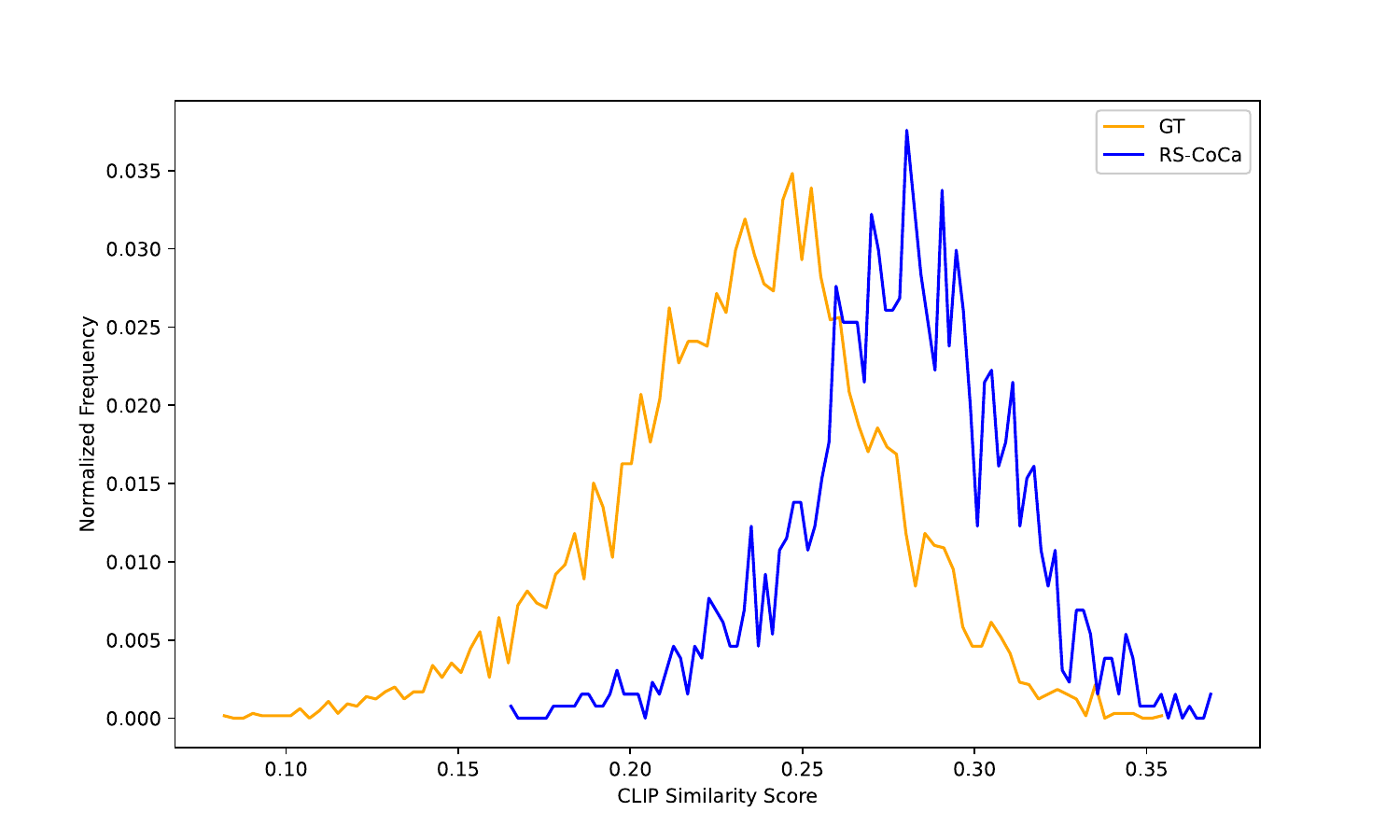}
  \caption{CLIP Similarity Score Distribution Between RS-CoCa Captions and GT on Benchmark Dataset }
  \label{fig:CoCaCLIPScoreDistribution}
\end{figure}

\begin{figure*}[htb]
  \centering
  \includegraphics[width=1\textwidth]{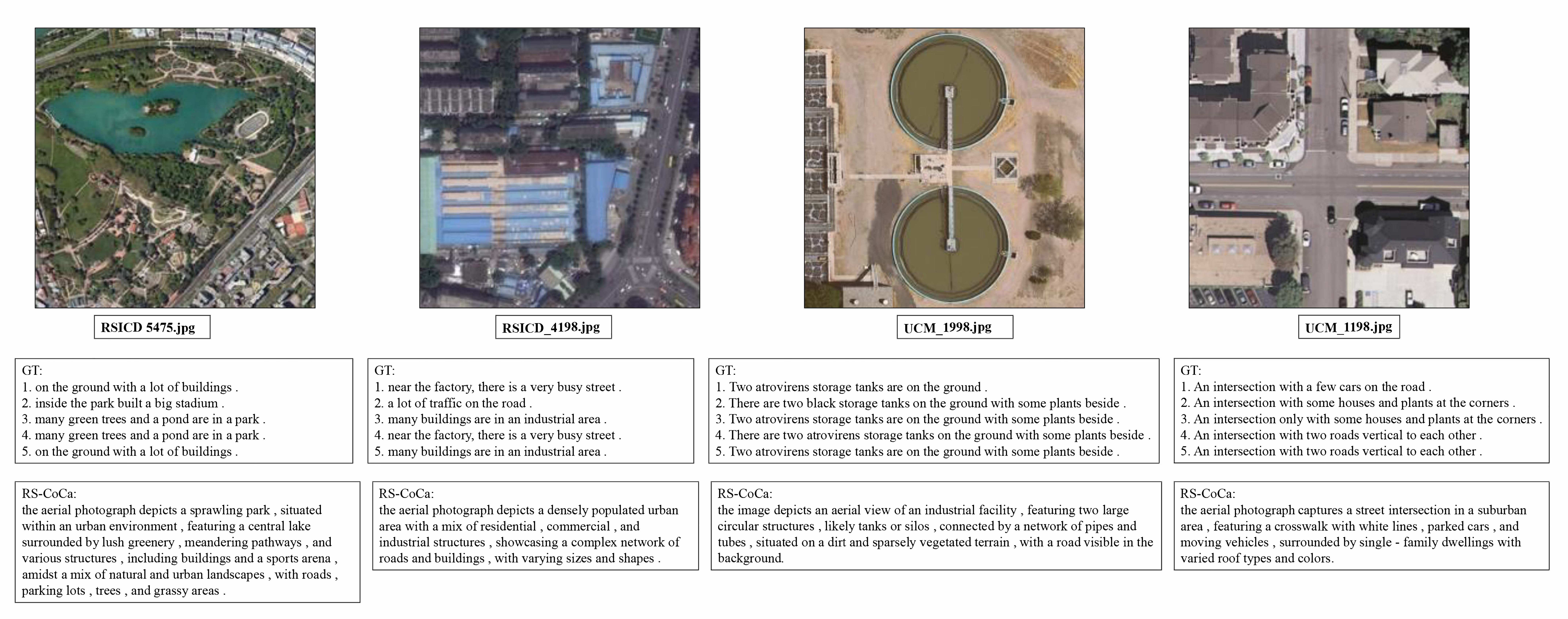}
  \caption{Sampled Captions generated by RS-CoCa. }
  \label{fig:RSCoCasamples}
\end{figure*}

\textbf{Qualitative Analysis.} As shown in Figure \ref{fig:RSCoCasamples}, the captions generated by RS-CoCa exhibit exceptionally high quality, comparable to or even surpassing human annotations. Manual evaluation confirms that RS-CoCa achieves annotation accuracy on par with human performance, while demonstrating higher image-caption similarity in CLIP similarity score distributions(Figure \ref{fig:CoCaCLIPScoreDistribution}). Additionally, the average length of RS-CoCa captions is 48.88 words, significantly exceeding the 10.34 words of human annotations, thereby providing much richer information. This comprehensive and precise description not only enhances the interpretation of RS images but also facilitates more effective image-text alignment during VLMs training.

\begin{table*}[h!]
\centering
\caption{Results of RS Captioning task. The best result is in \textbf{bold}.}
\begin{tabular}{l l c c c c c c c c}
\toprule
\textbf{Dataset} & \textbf{Method}    & \textbf{BLEU-1} & \textbf{BLEU-2} & \textbf{BLEU-3} & \textbf{BLEU-4} & \textbf{METEOR} & \textbf{ROUGE\_L} & \textbf{CIDEr} & \textbf{SPICE} \\ 
\midrule
\multirow{5}{*}{RSICD} 
                & BITA              & 0.774           & 0.665           & 0.577           & 0.504           & 0.420  & \textbf{0.717}   & \textbf{3.045} & 0.548 \\ 
                & RSGPT             & 0.703           & 0.542           & 0.440           & 0.368           & 0.301           & 0.533            & 1.029          & NA             \\ 
                & SkyEyeGPT         & \textbf{0.867}  & \textbf{0.767}  & \textbf{0.673}  & \textbf{0.600}  & 0.354           & 0.626            & 0.837          & NA             \\ 
                & RS-CapRet         & 0.741           & 0.622           & 0.529           & 0.455           & 0.376           & 0.649            & 2.605          & 0.484          \\ 
                & Ours              & 0.757           & 0.626           & 0.527           & 0.449           & \textbf{0.425}           & 0.708            & 2.517          & \textbf{0.555}          \\ 
\midrule
\multirow{5}{*}{UCM} 
                & BITA              & 0.889           & 0.831           & 0.773           & 0.719           & 0.469           & 0.838            & \textbf{3.845} & 0.549          \\ 
                & RSGPT             & 0.861           & 0.791           & 0.723           & 0.657           & 0.422           & 0.783            & 3.332          & NA             \\ 
                & SkyEyeGPT         & \textbf{0.907}  & \textbf{0.857}  & \textbf{0.816}  & \textbf{0.784}  & 0.462           & 0.795            & 2.368          & NA             \\ 
                & RS-CapRet         & 0.833           & 0.760           & 0.699           & 0.645           & 0.447           & 0.786            & 3.429          & 0.525          \\ 
                & Ours              & 0.890           & 0.834           & 0.783           & 0.734           & \textbf{0.487}  & \textbf{0.842}   & 3.701          & 0.539 \\ 
\bottomrule
\end{tabular}
\label{tab:Captioning}
\end{table*}

\textbf{Quantitative Analysis.} Table~\ref{tab:Captioning} illustrates the performance of various models on the RSICD and UCM datasets across multiple metrics. Notably, RS-CoCa achieves state-of-the-art or comparable results to the leading methods in semantic relevance and sentence diversity metrics, including METEOR, ROUGE\_L, CIDEr, and SPICE. This underscores RS-CoCa's capability to generate rich and diverse descriptive captions effectively. In contrast, SkyEyeGPT exhibits superior performance in BLEU metrics, which can be attributed to its extensive use of manually annotated training data. This reliance results in a high degree of similarity between the training and test data distributions, thereby enhancing its performance on BLEU scores. Despite this, RS-CoCa maintains strong competitiveness in BLEU metrics, demonstrating its robust overall performance.

\subsection{Ablation Study of Dataset Construction Method}

We chose CLIP for the dataset construction ablation studies primarily because of its simplicity, computational efficiency, and focus on contrastive learning, which allows for more interpretable evaluations of the impact of different factors on multimodal alignment performance. Compared to the more complex CoCa model, which emphasizes generative tasks, CLIP is better suited for such experiments. 
We conducted ablation studies using cross-modal retrieval and SeLo tasks, as they are multimodal tasks that more effectively evaluate CLIP's multimodal alignment capabilities and reveal the impact of multimodal data and strategies.

To intuitively reflect the CLIP model's overall retrieval performance, we use \textbf{3ret} to represent the RSITMD, RSICD, and UCM test sets. The RS-CLIP model's performance on 3ret is averaged, with Aver I2T R@1, Aver T2I R@1, and Aver Mean Recall representing the average values of the corresponding metrics across the three test sets. 

\subsubsection{Effectiveness of Different Description Methods}
Captions that only describe a portion of an image's visual elements can cause category ambiguity~\citep{patternnet,rsicd}, leading to unclear positive and negative pairs during training and reducing the model's recognition capabilities. To test this, we trained two models using the same images with different captions: one with partial descriptions extracted by an LLM (RS-CLIP-part) and another with comprehensive descriptions (RS-CLIP-full).

\begin{table}[htbp]
\centering
\caption{Comparison of Generation Methods on Retrieval and SeLo Tasks.
 The best result is in \textbf{bold}. Comp means Comprehensive. Avg means Average.}
\setlength{\tabcolsep}{2pt} 
\footnotesize 
\begin{tabular}{@{}lcccc|cccccc@{}} 
\toprule
\multirow{2}{*}{Method} & \multirow{2}{*}{Test} & \multicolumn{2}{c}{Avg Recall @1} & \multirow{2}{*}{\makecell{Avg \\ Mean Recall}} & \multicolumn{4}{c}{Semantic Localization} \\ 
\cline{3-4} \cline{6-9}
                       &                        & I2T & T2I &                         & ↑ Rsu  & ↓ Ras  & ↑ Rda  & ↑ Rmi  \\ 
\midrule
Partial & 3ret    & 12.64        & 10.47        & 32.49            & 0.7346 & 0.3125 & \textbf{0.6953} & 0.7083 \\ 
Comp & 3ret    & \textbf{17.78} & \textbf{13.18} & \textbf{37.40} & \textbf{0.7526} & \textbf{0.2805} & 0.6909 & \textbf{0.7256} \\

\bottomrule
\label{table:AbationOfDescriptionMethod}
\end{tabular}
\end{table}

As shown in Table \ref{table:AbationOfDescriptionMethod}, the model trained with fully descriptive captions significantly outperformed the partially descriptive model in retrieval performance. In the SeLo task, metrics Rsu, Ras, and Rmi were all notably higher for RS-CLIP-full. These results indicate that providing complete captions for each image is essential for enhancing model performance.

\subsubsection{Effectiveness of Different Data scale }

\begin{figure}[htbp]
  \centering
  \includegraphics[width=0.5\textwidth]{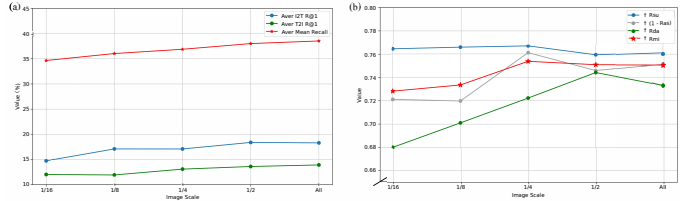}
  \caption{The Effectiveness of Different Data Scale.  }
  \label{fig:Ablation_Image_Scale}
\end{figure}

To investigate the impact of different numbers of training images on performance, we sampled 1/16, 1/8, 1/4, and 1/2 of the total images to train five models, named RS-CLIP-1 through RS-CLIP-5. The test results are shown in the Figure \ref{fig:Ablation_Image_Scale}.

As shown in Figure \ref{fig:Ablation_Image_Scale}(a), retrieval performance consistently improves with larger image datasets, indicating that more training data enhances retrieval capabilities. However, Figure \ref{fig:Ablation_Image_Scale}(b) reveals that SeLo task metrics do not steadily increase with dataset size. Metrics rise up to a 1/4 scale (approximately 50K images) but plateau or decline beyond that. Additionally, Rsu values remain consistently high across all scales. These results suggest that increasing the image scale does not necessarily improve semantic localization performance.

\subsubsection{Effectiveness of Different Caption Length}

\begin{figure}[htbp]
  \centering
  \includegraphics[width=0.5\textwidth]{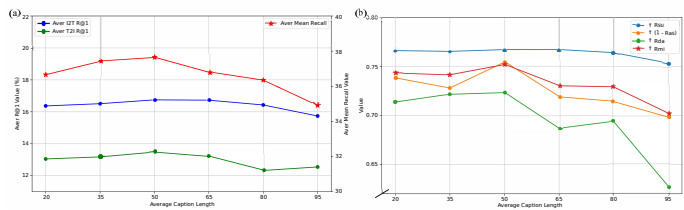}
  \caption{Effectiveness of Different Caption Length. }
  \label{fig:Ablation_Caption_length}
\end{figure}

\cite{DCI} demonstrated that condensing lengthy captions into concise, key visual descriptions can significantly enhance CLIP's performance. However, the effect of varying caption lengths on CLIP has not been thoroughly explored. In our experiment, we used LLMs to create captions of different lengths from the same descriptive information and trained CLIP models accordingly.

Figure \ref{fig:Ablation_Caption_length}(a) shows that retrieval performance improves as the average caption length increases up to around 50 words, and Figure \ref{fig:Ablation_Caption_length}(b) reveals a similar trend for the SeLo task. Beyond 50 words, longer captions lead to a decline in performance because many captions exceed CLIP's 77-token limit, causing text truncation and information loss. Therefore, while longer captions initially provide more information, surpassing the token limit reduces the model's effectiveness.

\subsubsection{Effectiveness of Different Caption Number per Image}
In the second stage, we instructed the LLM to generate five slightly different captions per image to introduce diversity and randomly selected one caption for analysis to manage computational costs. To explore how training with multiple captions affects CLIP's performance, we tested four image scales—10K, 50K, 100K, and 210K—and provided each image with 1 to 5 captions. We then trained the CLIP model to observe performance changes with varying numbers of captions.

\begin{figure}[htbp]
  \centering
  \includegraphics[width=0.45\textwidth]{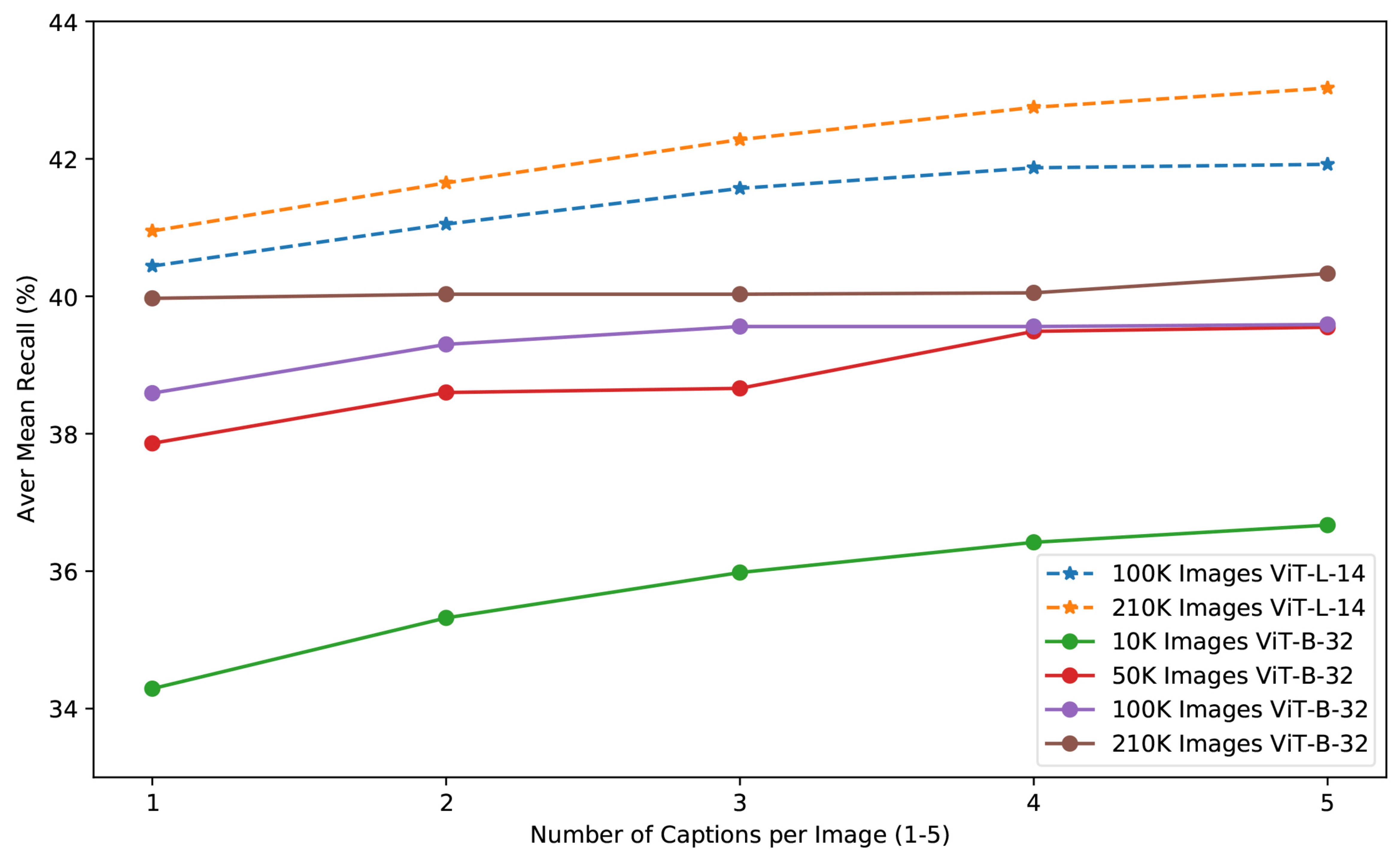}
  \caption{Effectiveness of Different Caption Numbers.  }
  \label{fig:Ablation_Image_Number}
\end{figure}

Figure \ref{fig:Ablation_Image_Number} shows that for smaller datasets (10K and 50K), the CLIP ViT-B-32 model's performance significantly improves as the number of captions increases, indicating that each caption adds unique information. However, for datasets larger than 100K, performance gains plateau despite more captions. This suggests that larger models are needed to fully utilize additional captions. Indeed, the CLIP ViT-L-14 model continued to show significant performance improvements on 100K and 210K image scales, highlighting that increasing the model size is necessary to leverage larger datasets effectively.

\subsubsection{Effectiveness of Different Sub-datasets}

To assess the impact of various sub-datasets, we divided HQRS-IT-210K into Drone*4, DET-10, CLS*8, and MSrgb*1 based on their sources. We trained CLIP on each sub-dataset and evaluated their performance to understand how different dataset types affect the model.

\begin{table}[htbp]
\centering
\caption{Comparison of Training Sub-datasets on RSCTIR and SeLo Tasks. Avg means Average.}
\setlength{\tabcolsep}{3pt} 
\scriptsize
\begin{tabular}{@{}lcccc|cccccc@{}}
\toprule
\multirow{2}{*}{Data (Number)} & \multirow{2}{*}{Test} & \multicolumn{2}{c}{Avg Recall @1} & \multirow{2}{*}{\makecell{Avg \\ Mean Recall}} & \multicolumn{4}{c}{Semantic Localization} \\ 
\cline{3-4} \cline{6-9}
                       &                        & I2T & T2I &                         & ↑ Rsu  & ↓ Ras  & ↑ Rda  & ↑ Rmi  \\ 
\midrule
All (210K)      & 3ret & 19.66 & 14.9  & 40.37            & 0.7635 & 0.2488 & 0.734  & 0.7518 \\
CLS*8 (93K)     & 3ret & 17.47 & 14.45 & 39.26            & 0.7665 & 0.2674 & 0.715  & 0.7417 \\
DET-10 (82K)    & 3ret & 16.13 & 13.48 & 36.11            & 0.7618 & 0.2957 & 0.6927 & 0.7244 \\
Drone*4 (33K)   & 3ret & 12.45 & 9.87  & 30.4             & 0.7475 & 0.2719 & 0.6522 & 0.7169 \\
MSrgb*1 (22K)     & 3ret & 12.34 & 10.56 & 31.26            & 0.7478 & 0.3044 & 0.6747 & 0.7113 \\
\midrule
CLIP-Baseline  & 3ret & 8.02  & 7.67  & 24.32            & 0.7188 & 0.3006 & 0.6992 & 0.7071 \\
\bottomrule
\label{tab:AblationofSub-Datasets}
\end{tabular}
\end{table}

Table \ref{tab:AblationofSub-Datasets} shows that CLS*8 and DET-10 significantly improve CLIP's RS retrieval capabilities, while Drone*4 and MSrgb*1 also enhance performance but to a lesser degree. This difference is mainly due to two factors: firstly, CLS*8 and DET-10 have 3-4 times more data than Drone*4 and MSrgb*1, leading to better training results. Secondly, the images in CLS*8 and DET-10 are more similar to the test set, whereas Drone*4 and MSrgb*1 have lower distribution consistency with the test data. However, this inconsistency does not indicate lower quality of Drone*4 and MSrgb*1 datasets. In the SeLo task, a similar trend is observed. Models trained with CLS*8 and DET-10 outperform those trained with Drone*4 and MSrgb*1 in the Rsu metric, indicating better recognition of target regions.

\begin{table}[htbp]
\centering
\caption{Comparison of Construction Strategies in RSCTIR and SeLo Tasks.
Baseline is the result of CLIP. R-M R means Rule-MLLM Relay Generation method. The best result is in \textbf{bold}. Avg means Avg. Gen means Generation. Kom means Kosmos2. Lva means Llava1.6.Mec emans Mechanism. Prt means Prompt.}
\setlength{\tabcolsep}{4pt} 
\scriptsize
\begin{tabular}{@{}lcccc|cccccc@{}} 
\toprule
\multirow{2}{*}{Model} & \multirow{2}{*}{Test} & \multicolumn{2}{c}{Avg Recall @1} & \multirow{2}{*}{\makecell{Avg \\ Mean Recall}} & \multicolumn{4}{c}{Semantic Localization} \\ 
\cline{3-4} \cline{6-9}
                       &                        & I2T & T2I &                         & ↑ Rsu  & ↓ Ras  & ↑ Rda  & ↑ Rmi  \\ 
\midrule
Baseline     & 3ret     & 8.02   & 7.67   & 24.32  & 0.7188 & 0.3006 & 0.6992 & 0.7071 \\
\midrule
R-M R         & 3ret     & 11.33  & 9.49   & 30.68  & 0.7125 & 0.3111 & 0.6833 & 0.6969 \\
Lva-Gen     & 3ret     & 14.4   & 12.75  & 34.80  & 0.7609 & 0.2890 & 0.6718 & 0.7212 \\
Kom-Gen   & 3ret     & 15.27  & 12.26  & 36.50  & 0.7414 & 0.2803 & 0.7063 & 0.7250 \\
\midrule
LLM-Prt1   & 3ret     & 18.75  & 13.82  & 38.46  & 0.7622 & 0.2765 & 0.7338 & 0.7415 \\
LLM-Prt2     & 3ret     & 18.66  & 13.96  & 39.97  & 0.7594 & 0.2748 & 0.6996 & 0.7324 \\
\midrule
Fuse Mec & 3ret & \textbf{19.66} & \textbf{14.90} & \textbf{40.37} & \textbf{0.7635} & \textbf{0.2488} & \textbf{0.7340} & \textbf{0.7518} \\
\bottomrule
\end{tabular}
\label{table:AblationOfStrategies}
\end{table}

\subsubsection{Effectiveness of Different Construction Strategies}

During dataset construction, we meticulously evaluated each strategy through independent experiments to validate their effectiveness. In the first stage, we generated image captions using three methods: \textbf{R-M Relay}, \textbf{Llava1.6-Generation}, and \textbf{Kosmos2-Generation}. In the second stage, we employed \textbf{LLM-Prompt1} and \textbf{LLM-Prompt2} to guide LLMs in summarizing the captions. The third stage utilized a \textbf{Fuse Mechanism} to combine the captions.

Table \ref{table:AblationOfStrategies} demonstrates that performance consistently improved at each stage of caption generation, ultimately achieving the best results. This highlights the effectiveness of our dataset construction methodology. Additionally, we found that captions generated by Kosmos2 outperformed those from Llava1.6 after training CLIP. The primary reason is that Llava's captions were significantly longer (approximately 150 words) compared to Kosmos2's (approximately 44 words), leading to substantial text truncation during CLIP training and resulting in poorer performance.

\subsubsection{Effectiveness of Different \(\alpha\) Values}

After the two caption generation stages, we found that mixing two types of LLM-generated captions increases the diversity of the training data, thereby enhancing CLIP's performance. To identify the optimal mixing ratio, we conducted comparative experiments where $\alpha = 1$ used only “prompt 2” generated captions. The results showed that an $\alpha$ of approximately 0.5 yielded the best retrieval and SeLo performance, outperforming models trained with a single caption type. This optimal ratio likely balances data diversity by incorporating both caption styles and acts as regularization, enabling the model to leverage the strengths of each style and improve its generalization ability.

\section{Conclusions}

We developed a two-stage RS image captioning method that integrates advanced LLMs and MLLMs, resulting in the HQRS-IT-210K dataset. This high-quality dataset supports the development of additional RS models, thereby further advancing progress within the research community. 

By fine-tuning CLIP and CoCa on this dataset, we created HQRS-CLIP and RS-CoCa. HQRS-CLIP is a robust discriminative VLFM that outperforms previous models across various RS applications using significantly less training data. RS-CoCa is a powerful generative model capable of generating captions for RS images that are comparable to, or even surpass, the quality of captions annotated by human experts. Additionally, our extensive ablation experiments provided key insights into the caption generation process, offering valuable guidance for future research.


\ifCLASSOPTIONcaptionsoff
  \newpage
\fi

\bibliographystyle{IEEEtran}
\bibliography{IEEEabrv,IEEEtran}
\end{document}